\documentclass[acmlarge]{acmart}
\acmJournal{IMWUT}
\AtBeginDocument{%
  }

\setcopyright{cc}
\setcctype{by}
\acmYear{2026} \acmVolume{10} \acmNumber{4} \acmArticle{246}
\acmMonth{12} \acmDOI{10.1145/3857944}

\usepackage{subcaption}
\usepackage{graphicx}
\usepackage{multirow} 
\usepackage{float}
\begin{document}

\title{From Packets to Patterns: Interpreting Encrypted Network Traffic as Longitudinal Behavioral Signals}

\author{Rameen Mahmood}
\orcid{0009-0001-0960-0817}
\affiliation{%
  \institution{New York University}
  \city{New York}
  \state{New York}
  \country{United States}}
\email{rameen.mahmood@nyu.edu}

\author{Omar El-Shahawy}
\orcid{0000-0003-3342-6377}
\affiliation{%
  \department{Department of Population Health}
  \institution{NYU Langone Health}
  \city{New York}
  \state{New York}
  \country{United States}}
\email{omar.elshahawy@nyulangone.org}

\author{Souptik Barua}
\orcid{0000-0002-1675-8874}
\affiliation{%
  \department{Division of Precision Medicine}
  \institution{NYU Grossman School of Medicine}
  \city{New York}
  \state{New York}
  \country{United States}}
\email{souptik.barua@nyulangone.org}

\author{Zachary Beattie}
\orcid{0000-0002-4844-7122}
\affiliation{%
  \department{Oregon Alzheimer’s Disease Research Center \& Center for Aging \& Technology (ORCATECH)}
  \institution{Oregon Health \& Science University}
  \city{Portland}
  \state{Oregon}
  \country{United States}}
\email{beattiez@ohsu.edu}

\author{Jeffrey Kaye}
\orcid{0000-0002-9971-3478}
\affiliation{%
  \department{Oregon Center for Aging \& Technology; Departments of Neurology \& Biomedical Engineering}
  \institution{Oregon Health \& Science University}
  \city{Portland}
  \state{Oregon}
  \country{United States}}
\email{kaye@ohsu.edu}

\author{Xuhai ``Orson'' Xu}
\orcid{0000-0001-5930-3899}
\affiliation{%
  \department{Department of Biomedical Informatics}
  \institution{Columbia University}
  \city{New York City}
  \state{New York}
  \country{United States}}
\email{xx2489@columbia.edu}

\author{Chao-Yi Wu}
\orcid{0000-0002-2187-6509}
\affiliation{%
  \department{Department of Neurology, Massachusetts General Hospital}
  \institution{Harvard Medical School}
  \city{Charlestown}
  \state{Massachusetts}
  \country{United States}}
\email{chwu3@mgh.harvard.edu}

\author{Danny Yuxing Huang}
\orcid{0000-0002-1794-6105}
\affiliation{%
  \institution{New York University}
  \city{New York}
  \state{New York}
  \country{United States}}
\email{dhuang@nyu.edu}

\renewcommand{\shortauthors}{Mahmood et al.}



\received{1 May 2026}
\received[revised]{1 August 2026}
\received[accepted]{15 September 2026}

\begin{abstract}
Human behavior is difficult to observe continuously at scale, yet it leaves measurable traces in everyday device use. We test whether encrypted smartphone network traffic---a ubiquitous, always-on, passive sensing modality---can passively capture behavioral patterns related to sleep, stress, and loneliness. We model shared behavioral structure using a transformer backbone with per-user adapters, allowing the model to represent both typical individual behavior and deviations from it. To make these representations interpretable, we apply a sparse autoencoder to extract behavioral features corresponding to distinct patterns of activity. We relate these features to sleep disturbance, stress, and loneliness using generalized estimating equations with Mundlak decomposition, separating between-person differences from within-person changes over time. We find that the three outcomes reflect distinct temporal structures: stress is primarily associated with stable between-person differences, loneliness with within-person variation, and sleep disturbance with a combination of both. Notably, these within-person dynamics are not captured by predefined network-traffic features, demonstrating the value of learned representations for longitudinal behavioral sensing. These results establish encrypted network traffic as a viable passive sensing modality, revealing interpretable behavioral dynamics—particularly deviations from an individual’s baseline—that are not visible in raw traffic features.
\end{abstract}

\maketitle
\section{Introduction}

Smartphones are deeply embedded in daily life: over 90\% of U.S. adults own one \cite{pew_mobile_2025}, and individuals spend several hours per day engaging with them \cite{harmonyhit_screen_time_2025,winbush2025smartphone} for communication, entertainment, and work \cite{gupta2011ubiquitous}. These interactions generate digital traces that form the foundation of smartphone-based digital phenotyping, the continuous measurement of human behavior using data from personal digital devices \cite{onnela2016harnessing,torous2016new}.

Prior work demonstrates that GPS traces are associated with depression \cite{canzian2015trajectories, saeb2015mobile}, accelerometry and social contact patterns with anxiety \cite{jacobson2020digital}, app usage and screen time with stress and sleep quality \cite{wang2014studentlife, nepal2024capturing}, and passive device signals more broadly with loneliness and social isolation \cite{doryab2019identifying}. Much of this work has been enabled by research platforms such as Beiwe \cite{onnela2021beiwe}, mindLAMP \cite{vaidyam2022enabling}, and Screenomics \cite{reeves2021screenomics}. Together, these platforms have enabled large-scale passive data collection and helped establish digital phenotyping as a widely used approach for behavioral and mental health research.

Despite this progress, smartphone-based sensing faces practical limitations. Many passive signals primarily capture \textit{when} a device is in use, offering limited insight into \textit{what the person is actually doing}. More granular signals, such as app-usage logs, can partially address this gap, but they rely on application- or OS-level instrumentation \cite{lee2022systematic, krieter2018analyzing}. This instrumentation is increasingly restricted by platform and privacy-policy changes \cite{slade2025current, harari2016using, schoedel2024mobile, schoedel2026person} and requires continual adaptation to evolving APIs \cite{mcdonnell2013empirical, mahmud2022android, wang2022aper}. As a result, longitudinal deployment becomes difficult due to reduced reliability over time \cite{bardram2022software}, increased engineering burden \cite{bardram2022software}, battery overhead that can affect participant compliance \cite{boonstra2018using,slade2025current}, and limited cross-platform scalability \cite{reeves2021screenomics,barnett2019intelligent,hossain2016challenges,schoedel2026person}. Alternative sensing modalities partially address these limitations, but introduce trade-offs in hardware requirements, deployment constraints, or scalability (Table~\ref{tab:modality_comparison}).

We explore encrypted network traffic as a complementary passive sensing modality that sidesteps these constraints by leveraging a ubiquitous, continuously available signal across homes and devices, enabling scalable sensing without additional infrastructure \cite{mahmood2026digital}. Unlike mobility or activity sensors, encrypted network traffic captures how individuals interact with digital services through observable metadata—such as timing, traffic volume, and destination domains\footnote{These signals are obtained from limited information exposed by DNS queries and the TLS handshake (e.g., domain names).}—without accessing message content. These behavioral traces provide a platform-agnostic proxy for online activity that can be observed continuously over time.

Prior work has begun to leverage smartphone network traffic for behavioral sensing. Mahmood et al.\ \cite{mahmood2026digital} established the feasibility of VPN-based passive monitoring in real-world deployments, demonstrating that longitudinal data collection is practical across both iOS and Android devices. Yue et al.\ \cite{yue2020automatic} extended this direction by classifying depression severity from session-level network features, producing participant-level predictions from smartphone WiFi metadata. However, both approaches produce cross-sectional outputs: a single label or score per participant derived from aggregate summaries, without modeling how behavior evolves relative to an individual's baseline. This limits their utility in longitudinal settings, where within-person change is often the primary signal of interest—for example, when monitoring disease progression or response to intervention \cite{hamaker2015critique,mohr2017personal}—because longitudinal behavioral trajectories are reduced to a single participant-level outcome.

We build on this infrastructure \cite{mahmood2026digital} to address the more fundamental question prior work leaves open: rather than classifying participants from aggregate summaries, can encrypted network traffic reveal how behavior evolves within individuals over time? To our knowledge, this is the first work to answer this question by learning interpretable longitudinal behavioral representations from encrypted smartphone network traffic.

To this end, we investigate three well-being dimensions—sleep disturbance, stress, and loneliness—spanning short-term physiological states and longer-term social experiences. Outcomes are measured using validated short-form instruments (PROMIS-SD, PSS-4, UCLA-3), administered weekly to 42 university students over 7 weeks. In parallel, we collect hourly encrypted network traffic features via a VPN. This longitudinal design enables analysis of within-person behavioral change relative to each individual's baseline, rather than relying solely on cross-sectional differences.

Our work has two aims: (i) to learn interpretable behavioral patterns from encrypted network traffic that generalize across individuals, and (ii) to characterize how these patterns relate to sleep, stress, and loneliness, distinguishing between-person differences from within-person changes over time.

We address both aims through a unified behavioral representation. As longitudinal behavioral sensing requires capturing both regularities shared across individuals and deviations relative to each person's baseline, our representation is built around a transformer with lightweight per-user adapters. The shared backbone learns temporal patterns common across users, while the adapters personalize representations to each individual's baseline, enabling detection of meaningful within-person behavioral change. To make these representations interpretable, we apply a sparse autoencoder that decomposes them into a small set of features, each corresponding to a distinct behavioral pattern (e.g., time-of-day activity or app-category usage). These features serve as the interpretable units of longitudinal analysis. We then relate changes in feature activation over time to well-being outcomes using generalized estimating equations with Mundlak decomposition, separating between-person differences from within-person change.

\paragraph{Key contributions.} Our contributions are threefold. First, we demonstrate that passive encrypted network traffic—without app-level access, OS instrumentation, or platform-specific APIs—can support interpretable longitudinal behavioral sensing. Second, we show empirically that learned behavioral representations recover longitudinal behavioral patterns that are not captured by predefined network-traffic summaries, enabling analysis of both stable between-person differences and within-person behavioral change. Third, applying this framework in a 7-week deployment of 42 enrolled university students—of whom 25 contributed sufficient longitudinal data for analysis—we find that the same learned features exhibit distinct patterns of behavioral variation across outcomes: stress is primarily associated with stable between-person differences, loneliness with week-to-week within-person variation, and sleep disturbance with both. Together, these findings establish encrypted network traffic as a practical, platform-agnostic modality for studying behavioral change over time.
\section{Related Work}
\label{sec:related-work}
\subsection{Sensing Modalities for Behavioral and Health Research} 

Prior work on behavioral and health sensing can be broadly divided into two classes of passive sensing modalities: on-device sensing (smartphone-based digital phenotyping) and environmental sensing (e.g., RF/CSI-based approaches). Table~\ref{tab:modality_comparison} summarizes the signals captured by each modality and their deployment characteristics.

Smartphone-based digital phenotyping has demonstrated that passive smartphone signals provide complementary windows into human behavior and mental health \cite{torous2015realizing}. GPS-derived mobility trajectories have been linked to depressive symptom severity \cite{canzian2015trajectories}, with location variance and phone usage patterns correlating with PHQ-9 scores across short-term observational deployments \cite{saeb2015mobile}. Communication logs and Bluetooth proximity have been used to infer loneliness and social isolation \cite{doryab2019identifying, qirtas2022loneliness}, while more comprehensive frameworks combining GPS, accelerometry, and app usage have jointly modeled mental health, resilience, and behavioral change over extended periods \cite{wang2014studentlife, nepal2024capturing, xuan2025unlocking}. Collectively, these studies demonstrate that rich behavioral inference depends on access to high-level smartphone signals, a dependency that underlies many of the deployment challenges discussed below.

Achieving this level of behavioral richness, however, requires application- and OS-level instrumentation that is increasingly difficult to sustain. Privacy policy changes on iOS and Android have progressively restricted access to communication logs, location data, and app usage signals that earlier work relied on \cite{slade2025current, schoedel2024mobile, schoedel2026person, harari2016using}, while mobile APIs evolve rapidly (on the order of $\sim$100 updates per month on Android alone), requiring continual platform-specific maintenance to keep sensing pipelines operational \cite{mcdonnell2013empirical, mahmud2022android, wang2022aper}. Even when supported, continuous background sensing incurs measurable battery overhead that can reduce participant compliance over time \cite{boonstra2018using, slade2025current}, and sensing pipelines may lose coverage entirely following device reboots or OS interruptions without reactivation \cite{bardram2022software, mohr2017personal}. Screenomics captures fine-grained screen interactions at high resolution \cite{reeves2021screenomics} but depends on platform-specific instrumentation, limiting scalability across heterogeneous device ecosystems and mixed-platform study populations \cite{barnett2019intelligent, hossain2016challenges, schoedel2026person}.

RF- and WiFi-based sensing has demonstrated that ambient wireless signals can be repurposed for passive physiological monitoring without requiring wearable devices. Prior work has recovered respiration, heart rate, and body motion from wireless reflections \cite{adib2015smart,zhao2017learning,hsu2017extracting}, enabling downstream applications in sleep assessment \cite{ma2019wifi}, stress monitoring \cite{ha2021wistress}, and neurological and health monitoring more broadly \cite{yang2022artificial,yousefi2017survey,liu2022monitoring}. While these approaches provide rich physiological measurements, they typically require fixed hardware deployments and controlled environments, limiting scalability in real-world settings.

These sensing modalities expose a fundamental trade-off. Smartphone-based sensing captures rich behavioral context but depends on application- and OS-level instrumentation, while RF/CSI-based approaches provide rich physiological measurements but require dedicated hardware deployments. Consequently, no existing modality simultaneously enables continuous, platform-agnostic behavioral sensing at scale, motivating complementary approaches based on encrypted network traffic.
\begin{table*}[t]
\centering
\small
\caption{Comparison of passive sensing modalities for behavioral and health research. Checkmarks indicate signals each modality can capture directly; partial ($\sim$) indicates limited or constrained capture; dashes indicate the signal is outside the modality's scope. The three modalities are largely complementary: they overlap in coarse presence and activity signals but differ in physiological versus behavioral-content dimensions.}
\label{tab:modality_comparison}
\begin{tabular}{lccc}
\toprule
\textbf{Signal / Capability} & \textbf{Phone sensing} & \textbf{CSI / RF sensing} & \textbf{Network traffic} \\
\midrule
\multicolumn{4}{l}{\textit{Physiological signals}} \\
 Breathing rate / heart rate & -- & \checkmark & -- \\
 Sleep timing / duration / wake events & \checkmark\textsuperscript{a} & \checkmark & \checkmark\textsuperscript{a} \\
 Sleep staging (REM, N1--N3) & -- & \checkmark & -- \\
 Gait / fine motor patterns & $\sim$\textsuperscript{b} & \checkmark & -- \\
\midrule
\multicolumn{4}{l}{\textit{Behavioral context}} \\
 Location / mobility (GPS, place visits) & \checkmark & -- & $\sim$\textsuperscript{c} \\
 Physical activity (steps, accelerometry) & \checkmark & $\sim$ & -- \\
 Presence at home / occupancy & \checkmark & \checkmark & \checkmark \\
 Communication patterns (calls, SMS) & $\sim$\textsuperscript{d} & -- & $\sim$\textsuperscript{e} \\
 Social proximity (Bluetooth) & \checkmark & -- & -- \\
 Device engagement (screen, unlocks) & \checkmark\textsuperscript{d} & -- & $\sim$\textsuperscript{f} \\
 Application-level activity (app categories, content type) & $\sim$\textsuperscript{d} & -- & \checkmark \\
 Online timing and duration & \checkmark & -- & \checkmark \\
\midrule
\multicolumn{4}{l}{\textit{Deployment characteristics}} \\
 Platform-agnostic (iOS + Android) & $\sim$\textsuperscript{d} & \checkmark & \checkmark \\
 Requires no dedicated hardware & \checkmark & -- & \checkmark \\
 Continuous background collection & $\sim$\textsuperscript{d} & \checkmark & \checkmark \\
 Battery overhead on user device & high & low & low \\
\bottomrule
\end{tabular}

{\footnotesize
\textsuperscript{a}Indirectly inferred from inactivity windows, accelerometer signals, or low network activity; not equivalent to polysomnography.
\textsuperscript{b}Coarse motion via accelerometer; not equivalent to fine-grained gait analysis.
\textsuperscript{c}Inferable indirectly from network endpoints (e.g., home network vs.\ cellular).
\textsuperscript{d}Restricted by OS-level privacy policies; SMS and call log access is unavailable on iOS and tightly restricted under Android Play Store policies \cite{slade2025current, vaidyam2022enabling}. 
\textsuperscript{e}Visible only as encrypted flows to messaging endpoints, without content.
\textsuperscript{f}Network activity correlates with device engagement but does not capture screen state directly.}
\end{table*}

\subsection{Network Traffic as a Behavioral Signal}

Encrypted network traffic has increasingly been explored as a source of behavioral information despite payload encryption. Observable metadata—including packet timing, flow volume, and destination hostnames—has been shown to support device identification \cite{miettinen2017iot,10.1145/3808674}, application inference \cite{taylor2016appscanner}, and characterization of in-home activity patterns \cite{hu2023behaviot}. Traffic-rate metadata from commodity smart home devices has further been shown to reveal sleep, motion, and device usage even under end-to-end encryption \cite{apthorpe2018keeping}. Collectively, these studies demonstrate that encrypted traffic retains meaningful behavioral information despite concealing application payloads.

Recent work has begun extending network- and infrastructure-level data from descriptive analysis toward behavioral inference. Initial efforts used enterprise WiFi logs to characterize behavioral patterns associated with academic performance \cite{kesheng2020data} and population-level shifts in app usage during COVID-19 lockdowns \cite{ukani2021locked}. More recent work has moved toward individual-level behavioral assessment: router-level traffic has been used to derive features associated with depression-related symptoms \cite{nef2025carenet}, while session-level and aggregate network features have been used for participant-level depression classification \cite{ware2018large,yue2020automatic}. Complementing these inference efforts, VPN-based passive monitoring has been shown to support continuous data collection across iOS and Android devices in real-world deployments \cite{mahmood2026digital}.

Despite these advances, two limitations remain. First, existing behavioral inference approaches formulate the problem as participant-level classification: prior work predicts a single depression label from session-level or aggregate network features \cite{ware2018large,yue2020automatic}, without modeling how behavior evolves relative to an individual's own baseline over time. This limits their ability to capture within-person behavioral change, which is often the primary signal of interest in longitudinal studies \cite{hamaker2015critique,mohr2017personal}. Second, these approaches rely on hand-engineered features—including session-level timing, aggregate traffic statistics, and hostname-derived summaries—leaving open whether learned representations can recover richer behavioral structure from the same encrypted signal, and whether that structure tracks meaningful within-person change over time. 
Taken together, these gaps raise a central question: can encrypted network traffic support longitudinal behavioral analysis while recovering interpretable patterns of within-person behavioral change?
\section{Methods}
\subsection{Overview}
\begin{figure*}[htbp]
    \centering
    \includegraphics[width=\textwidth]{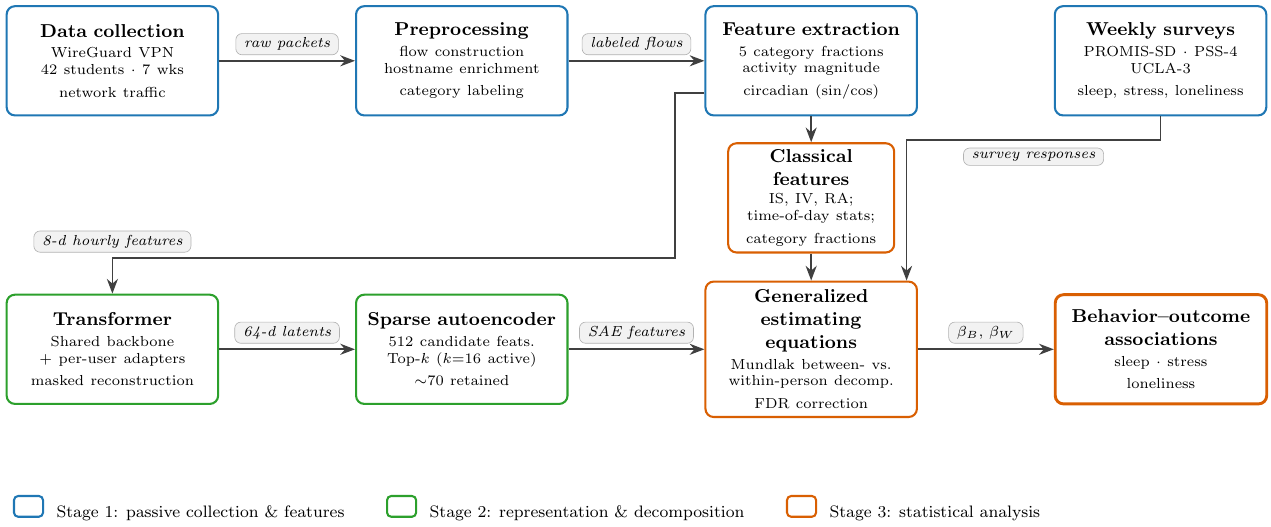}
    \Description{Three-stage pipeline diagram. Stage 1 shows data collection and feature construction from encrypted network traffic and surveys. Stage 2 shows representation learning with a transformer and per-user adapters, followed by a sparse autoencoder. Stage 3 shows statistical analysis relating learned and classical features to well-being outcomes.}
    \caption{Overview of the proposed framework. Stage 1: data collection and feature construction from encrypted network traffic and surveys. Stage 2: representation learning (transformer with per-user adapters) and sparse feature decomposition (SAE). Stage 3: statistical analysis relating learned and classical features to well-being outcomes using generalized estimating equations with Mundlak decomposition.}    \label{fig:pipeline}
\end{figure*}

Figure~\ref{fig:pipeline} summarizes our framework, which consists of three stages. Stage 1 collects encrypted smartphone network traffic via a WireGuard VPN and processes it into hourly behavioral feature vectors alongside weekly survey responses (Sec.~\ref{sec:data-collection}--Sec.~\ref{sec:feature-construction}). Stage 2 learns latent representations of behavior using a transformer with a shared backbone and per-user adapters, and decomposes these representations into interpretable sparse features using a sparse autoencoder (Sec.~\ref{sec:model-architecture}--Sec.~\ref{sec:sparse-autoencoder}). Stage 3 relates two predictor families---the learned sparse features and a set of classical hand-crafted features---to well-being outcomes (stress, loneliness, sleep disturbance) using generalized estimating equations with Mundlak decomposition to separate between-person and within-person effects (Sec.~\ref{sec:statistical-analysis}). 
\subsection{Study Design and Participants}

We conducted a 7-week observational study with 42 students at a major R1 university as part of a Big Data course. Students could select this study as one of several extra-credit options with comparable grading weight, ensuring voluntary participation and minimizing coercion. All participants provided informed consent prior to enrollment; details of the ethical approval, consent procedures, and data protection measures are provided in Sec.~\ref{sec:ethics}. Network traffic collection and weekly surveys were part of the course assignment, and the analysis presented here constitutes a secondary use of these data. No demographic information was collected because the study was designed to minimize the collection of potentially identifying information. Given the relatively small cohort and the longitudinal nature of the network traffic data, demographic attributes could substantially increase the risk of participant re-identification despite pseudonymization. Furthermore, because the study population comprised students at a single R1 university, it was relatively homogeneous with respect to age and life stage, and demographic variation was not central to the primary objective of evaluating within-person behavioral associations.

\paragraph{Inclusion criteria.}
Each participant-week included in the analysis satisfies four conditions: (i) the participant connected to the study VPN at least once; (ii) the VPN provided sufficient contiguous coverage to construct 48-hour input sequences for the model (Appendix~\ref{sec:week-filter}); (iii) the participant completed at least one weekly survey; and (iv) at least 48 hours of traffic precede the survey timestamp, providing the input window for behavioral inference.

\paragraph{Final cohort.}
Of the 42 enrolled participants, we excluded 1 who never connected to the VPN, 9 with insufficient contiguous coverage, 6 who could not be linked to weekly survey responses, and 1 without 48 hours of pre-survey traffic. The final cohort comprises 25 participants contributing 124 participant-week observations. Within-person analyses are restricted to the 24 participants with at least two observations. The full participant flow is reported in Appendix~\ref{app:consort}.

\subsection{Data Collection}
\label{sec:data-collection}
Participants installed a VPN profile on their personal smartphones by scanning a QR code that encodes a per-participant configuration \cite{donenfeld2017wireguard}. The VPN operates continuously across WiFi and cellular networks, requiring reactivation only after a device reboot. Participants received weekly email reminders to keep the VPN enabled and complete surveys. The overall setup and privacy model follow prior VPN-based passive monitoring work \cite{mahmood2026digital}.
When active, all device traffic is routed through a centrally operated VPN server, which we instrument using \texttt{tshark} \cite{wireshark_tshark_manual} to capture packets at the network interface. Each participant is assigned a unique local IP address at configuration time, enabling network traffic flows\footnote{We use the standard definition of a network flow as a source--destination IP pair together with packet-level statistics (e.g., timestamps, packet counts, and byte counts).} to be attributed to individual users despite multiplexing through a shared endpoint. This design supports fine-grained per-participant traffic separation while preserving end-to-end encryption.

\subsection{Data Processing and Feature Extraction}
We process captured traffic into flow-level records, where each flow is identified by a local VPN IP address (identifying the participant) and a remote IP address (identifying the external service). The local IP is used only for participant attribution and is not used to infer activity or service type. To improve interpretability, we map remote IP addresses to hostnames\footnote{Hostnames are obtained from DNS records and, when available, the Server Name Indication (SNI) field from TLS Client Hello packets.} and corresponding base domains. Traffic is first partitioned into intermediate 10-second intervals. Within each interval, and for each local--remote IP pair, we compute total upload and download packet counts and byte counts to form intermediate flow records. We selected a 10-second interval to preserve short-term temporal information during this intermediate processing step. These flow records are then aggregated into hourly behavioral features (Sec.~\ref{sec:hourly-aggregation}), which serve as the inputs to our behavioral models. Thus, the intermediate aggregation interval is independent of the temporal resolution used for behavioral modeling, and using a coarser interval (e.g., 60 seconds) would not affect the hourly behavioral features used in this work. As packet capture and intermediate aggregation are performed entirely on the centrally operated VPN server rather than on participants' devices (Sec. ~\ref{sec:data-collection}), the choice of aggregation interval affects only server-side processing and has no impact on device battery consumption, which is limited to the standard VPN tunnel overhead (<1\% \cite{mahmood2026digital}).

\subsection{Privacy and Ethical Considerations}
Data collection is limited to encrypted network metadata (e.g., timing, traffic volume, and destination domains), similar in scope to what an Internet Service Provider could observe, but without linkage to personally identifiable information. Consistent with prior work \cite{mahmood2026digital}, we do not collect payload content or app-level instrumentation, and all traffic is associated only with a study-specific identifier derived from the assigned VPN IP address. This design enables per-participant analysis while preserving end-to-end encryption.

\subsection{Survey and Outcome Measures}

Alongside passive collection of encrypted network traffic, participants completed weekly surveys measuring sleep disturbance, stress, and loneliness using validated short-form instruments (PROMIS-SD, PSS-4, and UCLA-3), each with well-established psychometric properties \cite{cella2010patient, cohen1983global, russell1996ucla}. Weekly survey completion was high (median 88\% relative to each participant's enrollment period), item-level missingness was negligible (0.2\%), and one duplicate submission was identified and removed prior to analysis. As a post hoc reliability check, we evaluated the internal consistency of each questionnaire—that is, whether its items produced similar responses among participants in our cohort. Cronbach's $\alpha$ ranged from 0.54 to 0.71 (Table~\ref{tab:survey-reliability}), indicating modest to acceptable internal consistency across the three questionnaires.

Survey responses were linked to traffic data via pseudonymous identifiers derived from the VPN configuration, enabling longitudinal analysis without revealing participant identity.

We focus on three well-being outcomes:

\begin{itemize}
\item \textbf{Sleep disturbance.} Measured using four items from the PROMIS Sleep Disturbance bank \cite{cella2010patient}, capturing sleep quality, refreshing sleep, sleep problems, and difficulty falling asleep. Items were coded so that higher scores indicate worse sleep and summed to form a composite score (range 4--20).

\item \textbf{Stress.} Assessed using the Perceived Stress Scale (PSS-4) \cite{cohen1983global}. Items were scored on a 1--5 scale, with positively worded items reverse-coded so that higher scores indicate greater perceived stress. Scores were summed to yield a total range of 4--20.\footnote{This differs from the canonical 0--4 scoring by a constant shift and does not affect correlations or standardized coefficients.}

\item \textbf{Loneliness.} Measured using the 3-item UCLA Loneliness Scale \cite{russell1996ucla}, capturing feelings of lacking companionship, feeling left out, and isolation. Higher scores indicate greater loneliness (range 3--9).
\end{itemize}

\subsection{Feature Extraction}

\subsubsection{Hourly Aggregation}
\label{sec:hourly-aggregation}
We aggregate flows into hourly bins for each participant, computing per-hour traffic statistics including upload and download bytes and flow counts. Timestamps are converted from UTC to the participant’s local timezone (US Eastern Time) so that hour-of-day reflects diurnal activity patterns. Hours with no observed traffic are not zero-filled, as this would conflate missing data with inactivity. We therefore retain only hours with at least one observed flow. Additional filtering to ensure contiguous weekly coverage is described in Appendix~\ref{sec:week-filter}.
\subsubsection{App-Category Mapping}
We map each network flow to a behavioral category (e.g., messaging, social media, video streaming) in two stages: hostname-to-application and application-to-category. Both mappings are defined using a manually curated dictionary; full details of the mapping pipeline are provided in Appendix~\ref{app:dictionary}.

\paragraph{Hostname to application.} 
We map each destination hostname (e.g., \texttt{youtube.com}) to an application using a manually curated dictionary of 231 hostnames covering 92 applications. The mapping is not one-to-one, as applications may use multiple domains (e.g., \texttt{whatsapp.com}, \texttt{whatsapp.net}).

\paragraph{Application to category.}
We group applications into 11 \emph{traffic categories} to account for all observed application traffic. Five are used as model inputs: four correspond to distinct types of user-driven activity (\textit{communication}, \textit{social media}, \textit{streaming}, and \textit{productivity}), while a fifth (\textit{system}) captures background operating-system traffic. Each application is assigned to a single primary category based on its dominant function (e.g., WhatsApp and Messenger as \textit{communication}, Facebook and Instagram as \textit{social media}).
\paragraph{Category selection.}
Of the 11 traffic categories, only five are used as model inputs. The remaining six are excluded for different reasons. \textit{CDN} and \textit{analytics} correspond primarily to shared infrastructure traffic that cannot be reliably attributed to user behavior, while the residual \textit{other} category contains unmapped hostnames retained only for completeness.
The remaining three behavioral categories (\textit{gaming}, \textit{browsing}, and \textit{shopping}) are excluded based on two predefined criteria designed to ensure stable longitudinal estimates: (i) sufficient participant coverage (non-zero weekly activity in $\geq$50\% of weeks for $\geq$40\% of participants) and (ii) sufficient traffic contribution ($\geq$1\% of total bytes).

\textit{Gaming} fails the coverage criterion. Although it accounts for 1.2\% of total traffic, only 32\% of participants generate any gaming traffic, and a single participant contributes 79\% of all gaming bytes. Including this category would therefore produce unstable population-level estimates dominated by a small number of participants.

\textit{Browsing} and \textit{shopping} fail the volume criterion despite being widely observed. Browsing is present in 95\% of participants but contributes only 0.7\% of total traffic, while shopping is observed in 93\% of participants yet contributes only 0.4\%. At these volumes, weekly category fractions become numerically unstable because small absolute changes in activity produce large relative fluctuations, making them unsuitable as longitudinal predictors.

Excluded traffic is retained when computing total traffic volume and therefore contributes to the denominator of every behavioral category fraction (Sec.~\ref{sec:feature-construction}), but not to any behavioral category numerator.

\paragraph{Coverage and validation.}
Overall, the dictionary covers approximately 77\% of total traffic volume; the remaining 23\% corresponds to a long tail of unmapped services that contribute only to the denominator when computing category-level fractions. As a validation check, we compare our application-to-category assignments for the five model-input categories against Apple App Store genres using the iTunes Search API \cite{apple_itunes_search_api}, finding no contradictions. Full validation details are provided in Appendix~\ref{app:dictionary}.

\subsubsection{Feature Construction}
\label{sec:feature-construction}
We construct an 8-dimensional feature vector per participant-hour, consisting of behavioral category fractions, activity magnitude, and circadian encoding.

\begin{itemize}
\item \textbf{Behavioral category fractions} (5 dimensions):  
For each category $c$, we compute
\[
\text{frac}_c =
\begin{cases}
\dfrac{\text{bytes}_c}{\text{total bytes}}, & \text{if total bytes} > 0,\\
0, & \text{if total bytes} = 0.
\end{cases}
\]
where $\text{bytes}_c$ is the total bidirectional traffic volume (upload + download) attributed to category $c$ within the hour,\footnote{Upload and download traffic are aggregated to capture total activity volume, as a single interaction may generate traffic in both directions.} and $\text{total bytes}$ is the total traffic across all categories, including unmapped traffic.
These features represent the distribution of activity across behavioral categories. 
\item \textbf{Activity magnitude} (1 dimension):  
We measure overall activity using the number of observed flows, which captures how frequently the device communicates with external services. To account for large differences across participants, flow counts are normalized within each user by converting them to percentile ranks over the full observation period and rescaling to $[0,1]$.\footnote{Percentiles are computed over the full observation period to preserve meaningful variation in activity levels; per-day normalization would compress differences and obscure inactive periods.}

\item \textbf{Circadian encoding} (2 dimensions):  
To represent time of day while preserving its cyclical structure, each 
hour $h \in \{0, 1, \ldots, 23\}$ is encoded as
\[
\left( \sin\left(\frac{2\pi h}{24}\right), \ \cos\left(\frac{2\pi h}{24}\right) \right).
\]
This encoding preserves temporal continuity across the midnight boundary and uniquely represents each hour.\footnote{Both sine and cosine are required to avoid ambiguity (e.g., 03:00 and 21:00 share the same sine value).}
\end{itemize}

\subsection{Model Architecture: Transformer with Per-User Adapters}
\label{sec:model-architecture}
\subsubsection{Modeling Rationale.} 
Passive network traffic exposes \emph{when} and \emph{how} a device is used, but not the underlying behavior directly. Individual observations (e.g., communication or streaming activity in a given hour) become interpretable only in context---both as part of recurring routines and as deviations from those routines over time. Our goal is therefore to learn representations that capture this longitudinal structure from hourly behavioral features.

Two properties of the problem shape the model design. First, interpreting behavior requires reasoning over temporal context: the meaning of an observation depends on surrounding hours (e.g., activity at 11am vs.\ 3am), and deviations are defined relative to recent and day-aligned patterns. We therefore use a transformer, whose self-attention mechanism allows each hour to attend to all others within a context window, capturing both local interactions and longer-range structure (e.g., recurring time-of-day patterns across days).

Second, behavioral baselines are user-specific: the same observable activity may reflect routine for one individual and deviation for another. A fully shared model would obscure these differences, while a fully per-user model would lack sufficient data to learn generalizable structure. We therefore use a shared transformer backbone with per-user residual adapters. The backbone captures patterns common across users, while lightweight per-user networks adjust representations to individual baselines, enabling detection of within-person deviations.

\subsubsection{Learning Objective.}
As no ground-truth behavioral labels exist at the hourly level, we train the model using a self-supervised \emph{masked reconstruction} objective, adapted from masked language modeling \cite{devlin2019bert} and its extension to multivariate time series \cite{zerveas2021transformer}. Given a 48-hour sequence, one hour is randomly masked and reconstructed from the remaining context using mean squared error (MSE) loss.
We use a 48-hour context window to ensure that each masked hour has a corresponding instance at the same time of day within the sequence. A 24-hour window lacks this property, as each hour appears only once. Longer windows provide additional redundant references without introducing new circadian structure, while reducing the number of valid training sequences (Appendix~\ref{sec:week-filter}). The 48-hour choice therefore balances temporal context and data availability, encouraging the model to learn patterns that recur across days rather than smoothing over adjacent hours. The 48-hour window defines only the temporal context used during representation learning; the longitudinal analysis is performed separately on repeated weekly observations (Sec.~\ref{sec:gee}).

\subsubsection{Model Input and Output.}
The model takes as input a 48-hour sequence of 8-dimensional feature vectors (5 category fractions, 1 activity magnitude, and 2 circadian dimensions; Sec.~\ref{sec:feature-construction}) and produces a 64-dimensional representation for each hour.
After training, these representations are passed to a sparse autoencoder (Sec.~\ref{sec:sparse-autoencoder}), which decomposes them into a small set of interpretable behavioral features for downstream statistical analysis.

\subsubsection{Architecture.} 
\label{sec:architecture}
The architecture comprises three components: a shared transformer backbone, per-user residual adapters, and a shared prediction head. Each 48-hour input window is processed through these components in sequence.

\begin{figure}[htbp]
    \centering
    \includegraphics[width=0.95\textwidth]{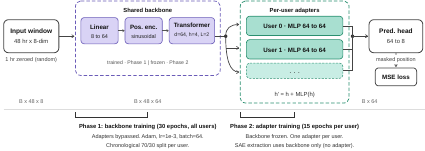}
    \Description{Architecture diagram of the two-phase training procedure, with purple boxes for the shared transformer backbone components and green boxes for the user-specific residual adapters.}
    \caption{Two-phase training procedure for the shared transformer backbone and per-user residual adapters. Purple boxes denote backbone components; green boxes denote user-specific adapters.}
    \label{fig:example}
\end{figure}

\paragraph{Shared backbone.} Each hourly feature vector is projected from 8 to 64 dimensions, combined with a sinusoidal positional encoding, and passed through a 2-layer transformer encoder ($d_{\text{model}} = 64$, 4 attention heads). Self-attention allows each hour’s representation to incorporate information from the full 48-hour context.

\paragraph{Per-user residual adapter.} Each user is assigned a 2-layer multilayer perceptron (MLP) that adapts the shared backbone representation via a residual update:
\[
h_{\text{adapted}} = h_{\text{backbone}} + \mathrm{MLP}(h_{\text{backbone}}).
\]
This formulation constrains the adapter to make small, targeted adjustments, preserving shared temporal structure while accounting for user-specific baselines. It also enables efficient personalization: adding a new user requires training only the adapter, without modifying the shared backbone.

\paragraph{Shared prediction head.} The prediction head maps the adapted 64-dimensional representation at the masked time step to an 8-dimensional reconstruction of the original hourly features. The head is shared across users, so all representations are decoded through the same mapping. As a result, personalization is handled entirely by the adapter, and representations remain comparable across users.

\subsubsection{Training Procedure.} We train the model in two phases. In Phase 1 (shared backbone training), the backbone and prediction head are trained jointly on data from all users for 30 epochs with adapters disabled. In Phase 2 (per-user adaptation), the backbone and prediction head are frozen, and each user’s adapter is trained independently for 15 epochs using only that user’s data.

\paragraph{Evaluation.} For each user, we split data chronologically into 70\% training and 30\% test windows, ensuring that all test windows occur strictly after training windows to prevent temporal leakage. Adapters are trained only for users with at least one valid training window under this split. All phases use the Adam optimizer (learning rate $10^{-3}$, batch size $64$, seed $42$); full hyperparameters are reported in Appendix Table~\ref{tab:hyperparams}.

\subsection{Sparse Autoencoder Decomposition}
\label{sec:sparse-autoencoder}
\subsubsection{Rationale.}
The transformer backbone learns latent representations that capture behavioral structure across users (Sec.~\ref{sec:architecture}), but these are not directly suitable for our goal of relating specific behavioral patterns to well-being outcomes. The 64-dimensional representations are dense and \emph{entangled}: individual dimensions mix multiple behavioral factors and do not correspond to interpretable patterns \cite{bricken2023towards, elhage2022toy}. As a result, effects cannot be attributed to specific behaviors.

We therefore apply a sparse autoencoder (SAE) to decompose these representations into a set of sparse, interpretable features, each capturing a distinct behavioral pattern. 

\subsubsection{SAE Architecture and Training.}
\label{sec:sae-architecture-training}
The SAE encodes each 64-dimensional latent vector $\mathbf{x}$ into a sparse activation $\mathbf{z}$ over a learned dictionary of 512 latent features. Rather than allowing all 512 features to contribute to each representation, the TopK operator retains only the $k$ largest activations. In our primary configuration, we set $k=16$, meaning that each latent vector activates at most 16 features (approximately 3\% of the dictionary). The encoded representation is then decoded to reconstruct the original latent:
\begin{align*}
\mathbf{z} &= \mathrm{TopK}\!\left(
\mathrm{ReLU}\!\left(\mathbf{W}_e\,(\mathbf{x} - \mathbf{b}_{\text{pre}})\right),\, k
\right), \\
\hat{\mathbf{x}} &= \mathbf{W}_d \mathbf{z} + \mathbf{b}_{\text{post}},
\end{align*}
where $\mathbf{W}_e$ and $\mathbf{W}_d$ are the encoder and decoder weight matrices, $\mathbf{b}_{\text{pre}}$ and $\mathbf{b}_{\text{post}}$ are learned bias vectors applied before encoding and after decoding, respectively, and $\hat{\mathbf{x}}$ denotes the reconstructed latent representation. By restricting each latent vector to at most $k$ active features, the SAE encourages individual latent features to specialize in distinct behavioral patterns rather than distributing information across many features, improving interpretability of the learned representation \cite{bricken2023towards,cunningham2023sparse}. We set $k=16$, corresponding to the 1--5\% active-feature regime commonly used in prior SAE work \cite{templeton2024scaling, bricken2023towards}, and fixed this value before downstream statistical analysis. To assess the sensitivity of this design choice, we repeated the complete feature-selection and validation pipeline for $k \in \{4,8,16,32,64\}$ (Table~\ref{tab:topk_sensitivity}). The sweep identified a stable mid-sparsity regime ($k=8$--16), with the primary configuration ($k=16$) reproducing the primary feature-selection pipeline exactly (70 user-general features, 18 FDR-significant pairs, and 14 robust associations). At lower values ($k=4$), fewer robust behavioral features were recovered, whereas at higher values ($k\geq32$), the learned decomposition became increasingly fragmented, reducing the number of robust associations recovered.

The SAE is trained to reconstruct the latent vectors using mean squared error (MSE) loss. We use Adam (learning rate $3 \times 10^{-4}$, cosine annealing learning-rate schedule, batch size $512$) for 500 epochs. Five users (randomly selected; seed 42) are held out exclusively for validation, and the checkpoint with the lowest validation reconstruction error is selected for downstream analysis. To stabilize training, decoder columns are re-normalized to unit $L_2$ norm after each update.

\subsubsection{Feature Selection and Interpretation}
\label{sec:user-diversity-filter}

We select features for analysis based on two criteria: (i) cross-user generality, ensuring that features reflect behavior shared across participants, and (ii) interpretability, ensuring that features can be assigned clear behavioral meaning.

\paragraph{Cross-user generality filter.} 
For each feature, we identify the 50 windows with the highest activation and count the number of distinct participants represented. Features whose top activations span fewer than 40\% of participants are removed, as they likely capture user-specific behavior rather than population-level patterns. This yields 70 candidate features.

\paragraph{Feature labeling.} 
To interpret each feature, we examine its top-100 activating windows and compute the average value of each behavioral category fraction (Sec.~\ref{sec:feature-construction}). Categories are labeled \emph{high} if their average exceeds $1.5\times$ the population mean and \emph{low} if it falls below $0.7\times$; values between these thresholds are left unlabeled. The \textit{system} category is excluded from labeling, as it reflects background OS traffic rather than user-driven activity.

To assess the sensitivity of these thresholds, we repeated the labeling procedure across a 7×5 grid of high and low cutoff values (35 threshold pairs; Appendix~\ref{app:threshold-sens}, Table~\ref{tab:threshold-sens}). Across all threshold combinations, no behavioral category ever switched directly between \textit{high} and \textit{low}; changes occurred only between a labeled state and \textit{neutral}. Moreover, at least 83\% of individual label assignments were preserved across the threshold grid (median agreement: 95\%). Together, these results indicate that the behavioral interpretations are robust to reasonable variations in the labeling thresholds.

We then characterize each feature’s temporal pattern by identifying its most frequent activation hours and grouping them into time-of-day bands (e.g., night, morning, midday, afternoon, evening). This yields an initial descriptive label combining behavioral categories with the dominant time-of-day band (e.g., \textit{high communication, low streaming, midday}). Categories joined with ``+'' indicate co-occurrence (e.g., \textit{low streaming+social, night}), while features without category labels receive time-only labels (e.g., \textit{late-night activity (11pm--1am)}).
We refine these labels by examining the dominant foreground hostnames in the same top-activating windows, revealing the apps and content types driving each feature (e.g., messaging platforms, streaming services, work tools). The resulting labels are reported in Table~\ref{tab:feature_profiles}, with full per-feature foreground composition in Appendix Table~\ref{tab:forensic_labels}.

\paragraph{Label validation.}
We validate feature labels through a perturbation test. For each feature, we increase its activation by $5\times$ its empirical standard deviation while holding all others fixed, then pass the modified vector through the SAE decoder and prediction head to reconstruct the corresponding behavioral feature vector. Features labeled \emph{high} in a category increase that category’s fraction under perturbation, while \emph{low} features decrease it; across all retained features, these directional effects match the assigned labels.

\subsection{Statistical Analysis}
\label{sec:statistical-analysis}

We evaluate how SAE features (Sec. \ref{sec:sparse-autoencoder}) are associated with weekly self-reported sleep disturbance, stress, and loneliness. For each participant-week, each SAE feature is averaged across the seven days preceding the survey, yielding one feature vector per observation.

\subsubsection{Generalized Estimating Equations with Mundlak Decomposition}
\label{sec:gee}

Our analysis seeks to answer two complementary questions: (1) do individuals with different behavioral patterns report different levels of well-being, and (2) do changes in an individual's behavior over time coincide with changes in their own well-being? These correspond to between-person and within-person associations, respectively. To estimate both simultaneously, we use the Mundlak decomposition \cite{mundlak1978pooling}, which represents each time-varying predictor as two variables: a participant-specific mean that captures between-person differences, and deviations from that mean that capture within-person changes over time.

This decomposition is well suited to our setting for three reasons. First, a standard GEE without decomposition estimates a single population-average association, conflating between-person differences with within-person change \cite{liang1986longitudinal,neuhaus1998between}. Second, fixed-effects models estimate only within-person associations by absorbing all stable between-person variation, preventing analysis of persistent individual differences \cite{allison2009fixed,curran2011disaggregation}. Third, conventional random-effects models estimate both between-person and within-person effects, but assume that stable differences between participants do not bias these estimates \cite{hausman1978specification,bell2015explaining}. In behavioral studies, this assumption may be unrealistic because participants differ in stable characteristics that can influence both their digital behavior and their well-being. The Mundlak decomposition addresses this issue by explicitly separating each predictor into a participant-specific mean and deviations from that mean \cite{mundlak1978pooling,bell2015explaining}, allowing between-person differences and within-person changes to be estimated separately. Specifically, we decompose each predictor as
\begin{align}
    X_{it}^{\text{between}} &= \frac{\bar{X}_i - \bar{X}}{\sigma_X}, 
    & X_{it}^{\text{within}} &= \frac{X_{it} - \bar{X}_i}{\sigma_X},
\end{align}
where $\bar{X}_i$ is participant $i$'s mean predictor value across all observed weeks, $\bar{X}$ is the grand mean across all participants and weeks, and $\sigma_X$ is the pooled standard deviation of $X_{it}$. Thus, $X_{it}^{\text{between}}$ represents whether a participant typically exhibits higher or lower values of the predictor than the cohort average, whereas $X_{it}^{\text{within}}$ represents whether the participant's value in a given week is higher or lower than their own typical level. Both components are standardized by $\sigma_X$, making $\beta_B$ and $\beta_W$ directly comparable. We fit a separate univariate GEE for each (predictor, outcome) pair:
\begin{equation}
    Y_{it} = \beta_0 + \beta_B X_{it}^{\text{between}} + \beta_W X_{it}^{\text{within}} + \epsilon_{it},
\end{equation}
where $i$ indexes participants, $t$ indexes weeks, and $Y_{it}$ is the self-reported outcome for participant $i$ at week $t$. The intercept is denoted by $\beta_0$. The between-person coefficient, $\beta_B$, measures whether participants who typically exhibit higher values of a behavioral feature also tend to report higher (or lower) outcome scores than other participants. The within-person coefficient, $\beta_W$, measures whether weeks in which a participant exhibits more (or less) of a behavioral feature than is typical for that participant are also weeks in which they report higher (or lower) outcome scores.

As all predictors are standardized before modeling, the estimated coefficients are directly comparable across features. Specifically, each coefficient represents the expected change in the outcome associated with a one-standard-deviation increase in the corresponding behavioral feature. Positive coefficients indicate that greater feature activation is associated with higher outcome scores, whereas negative coefficients indicate that greater feature activation is associated with lower outcome scores. Statistical significance is determined using FDR-adjusted $q$-values.

We fit each model using GEE with an exchangeable working correlation structure, which accounts for the repeated weekly observations contributed by each participant while assuming equal correlation among observations from the same participant. Robust (sandwich) standard errors are reported to ensure valid statistical inference even if this working correlation structure is misspecified. Each SAE feature is evaluated in a separate univariate model to isolate its association with the outcome and avoid multicollinearity among features.

\subsubsection{Predictor Families}
We apply the GEE–Mundlak analysis to two families of predictors: learned SAE features and classical hand-crafted features.
The \emph{SAE features}—the filtered, interpretable features identified in Sec.~\ref{sec:sparse-autoencoder} and aggregated to weekly means—are the primary focus. These features are learned from data rather than pre-specified, capturing combinations of activity patterns and temporal structure that emerge during training.

The \emph{classical features} serve as a baseline and include standard measures from the digital phenotyping and circadian rhythm literature, including interdaily stability (IS), which measures how consistent activity patterns are from one day to the next; intradaily variability (IV), which measures how fragmented activity is within a day; and relative amplitude (RA), which measures the contrast between the most active and least active periods of the day \cite{van1999bright, witting1990alterations, gonccalves2015fresh}. We also include time-of-day activity statistics (e.g., morning, afternoon, evening, and night means; activity centroid; active span), weekday–weekend differences, and the five behavioral category fractions (Sec.~\ref{sec:feature-construction}). All features are computed from the same per-user rank-normalized hourly flow count (\texttt{flow\_count\_pct}) rather than raw counts, ensuring that rest–activity metrics reflect within-person temporal structure rather than differences in overall traffic volume; full mathematical definitions and formulas are provided in Appendix~\ref{app:classical-features}.

\subsubsection{Validation and Robustness Checks}
\label{sec:validation}

We validate the identified feature--outcome associations in three stages.

\paragraph{(1) Statistical significance:}

To control for multiple hypothesis testing, we apply the Benjamini--Hochberg false discovery rate (FDR) correction \cite{benjamini1995controlling} separately to the between-person ($\beta_B$) and within-person ($\beta_W$) coefficients, which capture between-person differences and within-person behavioral changes, respectively. Within each effect type, $p$-values are pooled across all feature--outcome pairs before correction. A feature--outcome pair is considered statistically significant if either its between-person or within-person association satisfies $q < 0.05$.

\paragraph{(2) Practical robustness:}

Statistical significance alone does not necessarily imply that an association captures a reliable or behaviorally meaningful signal. We therefore evaluate each FDR-significant pair using two complementary criteria: (i) \emph{predictive contribution}, measured as the relative reduction in RMSE compared with an intercept-only model, and (ii) \emph{directional consistency}, defined as the proportion of observations for which the model correctly predicts whether the outcome lies above or below the overall mean. Based on these criteria, each pair is classified as \textit{Robust} ($\Delta\mathrm{RMSE}\% > 0.5\%$ and sign consistency $> 60\%$), \textit{Unstable} ($\Delta\mathrm{RMSE}\% > 0.5\%$ only), \textit{Redundant} (sign consistency $> 60\%$ only), or \textit{Noise} (neither criterion met). We report only \textit{Robust} pairs as the primary findings; complete results for all FDR-significant pairs are provided in Appendix Table~\ref{tab:ablation_full}.

\paragraph{(3) Stability:}

Finally, we evaluate whether the identified \textit{Robust} associations remain consistent under reasonable variations of the analysis procedure. Specifically, we perform (i) leave-one-user-out (LOUO) resampling, in which the models are refit while excluding one participant at a time, and (ii) threshold sensitivity analyses, in which the $\Delta\mathrm{RMSE}\%$ and sign-consistency thresholds are systematically varied. Three feature--outcome associations remain \textit{Robust} across every threshold combination evaluated, indicating that the principal findings are not sensitive to the specific robustness criteria. Full threshold-sensitivity and per-feature LOUO results are provided in Appendix~\ref{sec:robustness-sensitivity-analyses}.
\section{Results}

\subsection{Participant Characteristics and Analysis Sample}

The analysis sample includes 25 of the 42 enrolled participants, contributing 124 participant-week observations. Each participant-week observation consists of one weekly survey paired with the preceding week's network traffic and serves as the unit of analysis throughout the study. Participants contributed between 2 and 9 participant-weeks (median = 6, IQR 4--6), reflecting differences in enrollment timing and periods of continuous VPN coverage.

\begin{figure*}[htbp]
\centering

\begin{subfigure}{\textwidth}
    \centering
    \includegraphics[width=\textwidth]{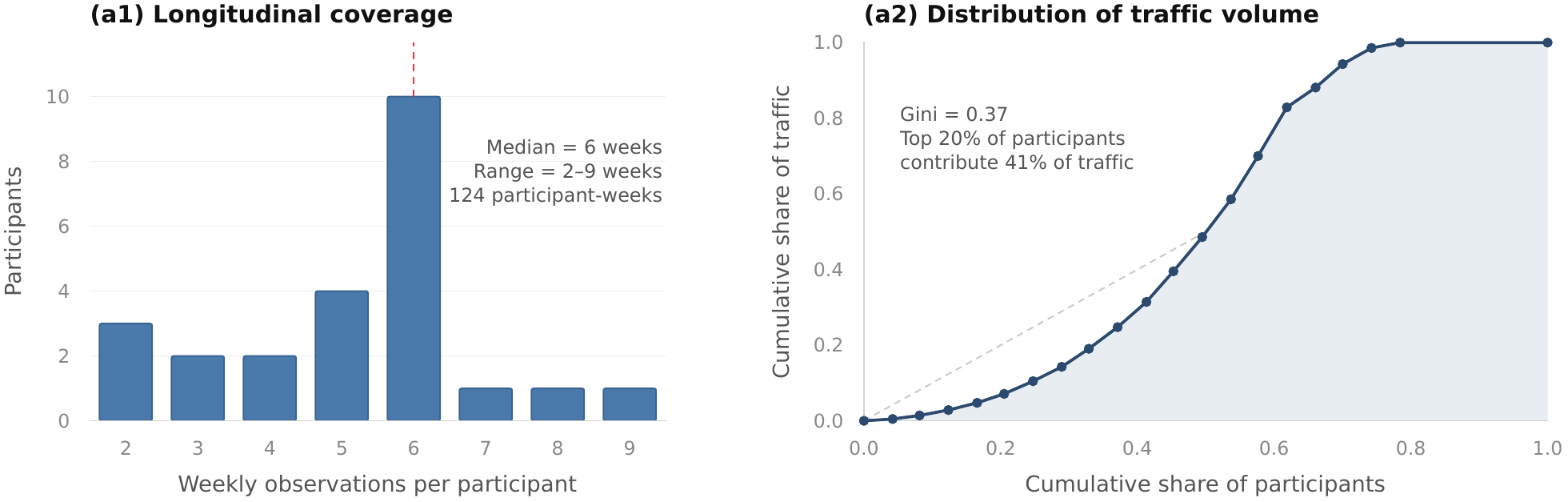}
    \Description{Two panels. The first is a bar chart of the number of participant-week observations contributed by each participant. The second is a Lorenz curve of cumulative network traffic volume across the 25 participants.}
    \caption{Participant coverage and traffic contribution.}
    \label{fig:participant_overview}
\end{subfigure}

\begin{subfigure}{\textwidth}
    \centering
    \includegraphics[width=\textwidth]{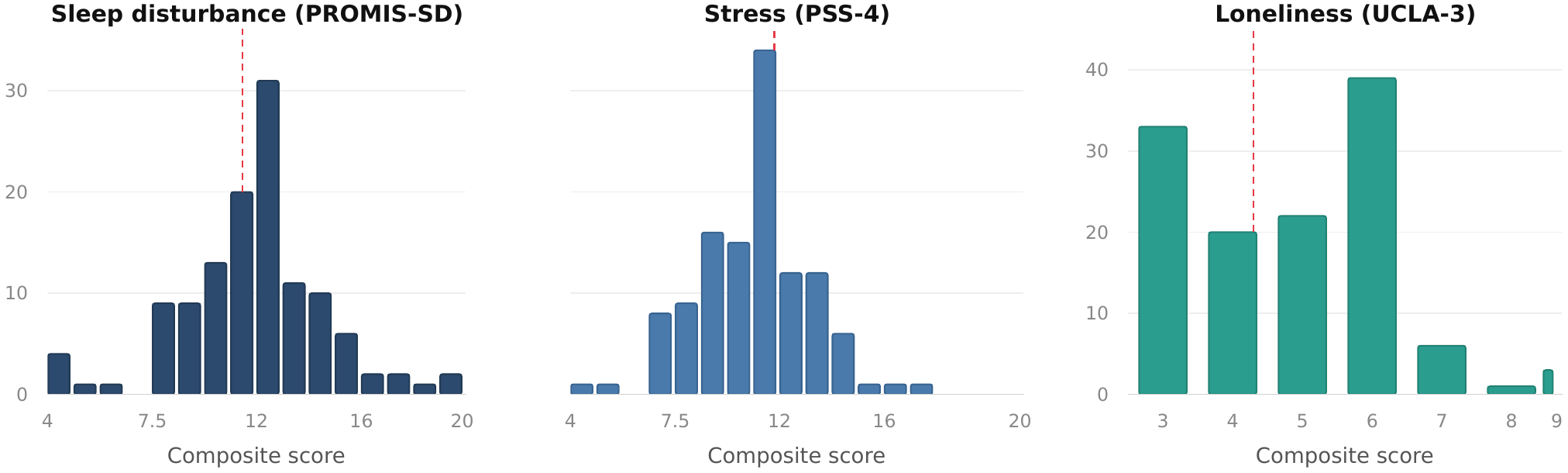}
    \Description{Three distribution plots showing weekly sleep disturbance, stress, and loneliness scores across the participant-week observations.}
    \caption{Distribution of weekly outcome measures.}
    \label{fig:outcome_distributions}
\end{subfigure}

\caption{Overview of the analysis cohort. (a) Number of participant-week observations contributed by each participant, where each observation consists of one weekly survey paired with the preceding week's network traffic (a1), and the cumulative distribution (Lorenz curve) of total network traffic volume across the 25 participants (a2). (b) Distributions of weekly sleep disturbance (PROMIS-SD), stress (PSS-4), and loneliness (UCLA-3) scores across the 124 participant-week observations included in the analysis. Dashed red lines indicate sample means.}
\label{fig:cohort_overview}
\end{figure*}
Figure~\ref{fig:cohort_overview} summarizes the characteristics of the analysis cohort. Panel (a1) shows the number of participant-week observations contributed by each participant, illustrating that most participants contributed multiple weekly observations (median = 6, range = 2--9). Panel (a2) shows the cumulative distribution of total network traffic volume across participants using a Lorenz curve. Although traffic volume varied across participants, the distribution was only moderately unequal (Gini = 0.37), with the top 20\% of participants accounting for 41\% of the total observed traffic, indicating that the dataset was not dominated by a small number of high-traffic participants. Panel (b) shows the distributions of the three weekly outcome measures across the 124 participant-week observations. Sleep disturbance, stress, and loneliness exhibited substantial variability, with means of 11.3 (SD = 3.0), 11.4 (SD = 2.4), and 4.9 (SD = 1.4), respectively. The outcome measures were modestly correlated (sleep disturbance--stress $r = 0.48$; stress--loneliness $r = 0.24$; sleep disturbance--loneliness $r = 0.14$), suggesting that they capture related but distinct aspects of well-being.
\subsection{Architecture Validation}
\subsubsection{Backbone: Decoding of Behavioral Metrics from Transformer Latents}
\label{sec:recovery-behavioral-metrics-transformer-latents}
Before relating SAE-derived features to well-being outcomes, we first assess whether the transformer backbone encodes meaningful behavioral structure. If basic behavioral signals are not present in the latent representation, downstream interpretation would be unreliable.

We evaluate this by predicting a set of known behavioral metrics from the latent representations, including time-of-day activity statistics (morning, evening, and night means; night-to-morning ratio), activity timing (activity centroid; active span), weekly structure (weekday–weekend difference), and circadian rhythm measures (interdaily stability, intradaily variability, relative amplitude). For each participant-week, we summarize the hourly latent vectors by their mean and standard deviation, then fit linear models using leave-one-subject-out cross-validation ($N=25$ folds).

\begin{table}[!htbp]
\centering
\caption{Backbone probing results. Each row reports recovery of a behavioral metric from transformer latent representations, summarized per participant-week by mean and standard deviation, under leave-one-subject-out cross-validation ($N=25$). \textbf{$r$:} Pearson correlation between predicted and true values (directional agreement). \textbf{$R^2$:} out-of-sample variance explained relative to the grand mean ($R^2 < 0$ indicates worse-than-mean prediction). \textbf{Encoded?:} \checkmark denotes statistically significant correlation ($p < 0.05$). Bold rows highlight metrics with the strongest recovery.}
\label{tab:probes}
\footnotesize
\begin{tabular}{llrrrl}
\toprule
\textbf{Category} & \textbf{Variable} & $r$ & $R^2$ & $p$ & \textbf{Encoded?} \\
\midrule
\multirow{3}{*}{Circadian}
  & Interdaily stability (IS) & 0.47 & 0.21 & 0.019 & \checkmark \\
  & Intradaily variability (IV) & 0.44 & 0.19 & 0.028 & \checkmark \\
  & Relative amplitude (RA) & 0.12 & $-$0.09 & 0.583 & \texttimes \\
\midrule
\multirow{2}{*}{Timing}
  & Activity centroid & 0.58 & 0.32 & 0.002 & \checkmark \\
  & Active span (hours) & 0.37 & 0.12 & 0.070 & \texttimes \\
\midrule
Weekly
  & Weekday--weekend difference & 0.59 & 0.34 & 0.002 & \checkmark \\
\midrule
\multirow{4}{*}{Time-of-day}
  & Morning activity & \textbf{0.75} & \textbf{0.55} & $<$0.001 & \checkmark \\
  & Night-to-morning ratio & \textbf{0.74} & \textbf{0.55} & $<$0.001 & \checkmark \\
  & Evening activity & \textbf{0.71} & \textbf{0.50} & $<$0.001 & \checkmark \\
  & Night activity & 0.59 & 0.35 & 0.002 & \checkmark \\
\bottomrule
\end{tabular}
\end{table}

Performance is evaluated using Pearson correlation $r$ (directional agreement) and out-of-sample $R^2$ (variance explained relative to the mean). A metric is considered \emph{recoverable} if $r$ is significantly positive ($p < 0.05$; checkmark in Table~\ref{tab:probes}).

The latent representations strongly encode time-of-day activity patterns, including morning activity ($r=0.75$, $p<0.001$), night-to-morning ratio ($r=0.74$, $p<0.001$), and evening activity ($r=0.71$, $p<0.001$). Weekly structure is also captured (weekday–weekend difference: $r=0.59$), along with circadian regularity measures such as interdaily stability ($r=0.47$) and intradaily variability ($r=0.44$).
In contrast, relative amplitude (RA) is not reliably decoded ($r=0.12$, $p=0.58$), as it depends on absolute activity levels removed by per-user normalization (Sec.~\ref{sec:feature-construction}). This confirms that the representation does not recover information excluded from the inputs, while still capturing meaningful structure in the remaining signals.
Together, these results show that the transformer latents encode key behavioral signals—including activity timing, weekly structure, and circadian regularity—providing a reliable foundation for subsequent SAE-based interpretation.

Furthermore, we assess robustness to participant-level variation. Per-user adapter deltas are nearly orthogonal across users (mean cosine similarity $0.02$), indicating that adapters capture user-specific adjustments rather than shared structure. Combined with the SAE cross-user generality filter (Sec.~\ref{sec:user-diversity-filter}) and leave-one-user-out resampling (Sec.~\ref{sec:validation}), this reduces the influence of any single participant on the reported results. Full analysis is provided in Appendix~\ref{app:adapter-specialization}.

\subsection{SAE-Derived Behavioral Features}
\label{sec:sae-dervied-behavioral-features}
Having validated the transformer backbone (Sec.~\ref{sec:recovery-behavioral-metrics-transformer-latents}), we apply a sparse autoencoder (SAE) to decompose the latent representations into sparse, interpretable behavioral features. Applying the SAE to 30{,}848 hourly latent vectors yields 337 active features (out of 512), of which 70 remain after the cross-user generality filter (Sec.~\ref{sec:user-diversity-filter}).

\begin{figure}[htbp!]
    \centering
    \includegraphics[width=\textwidth]{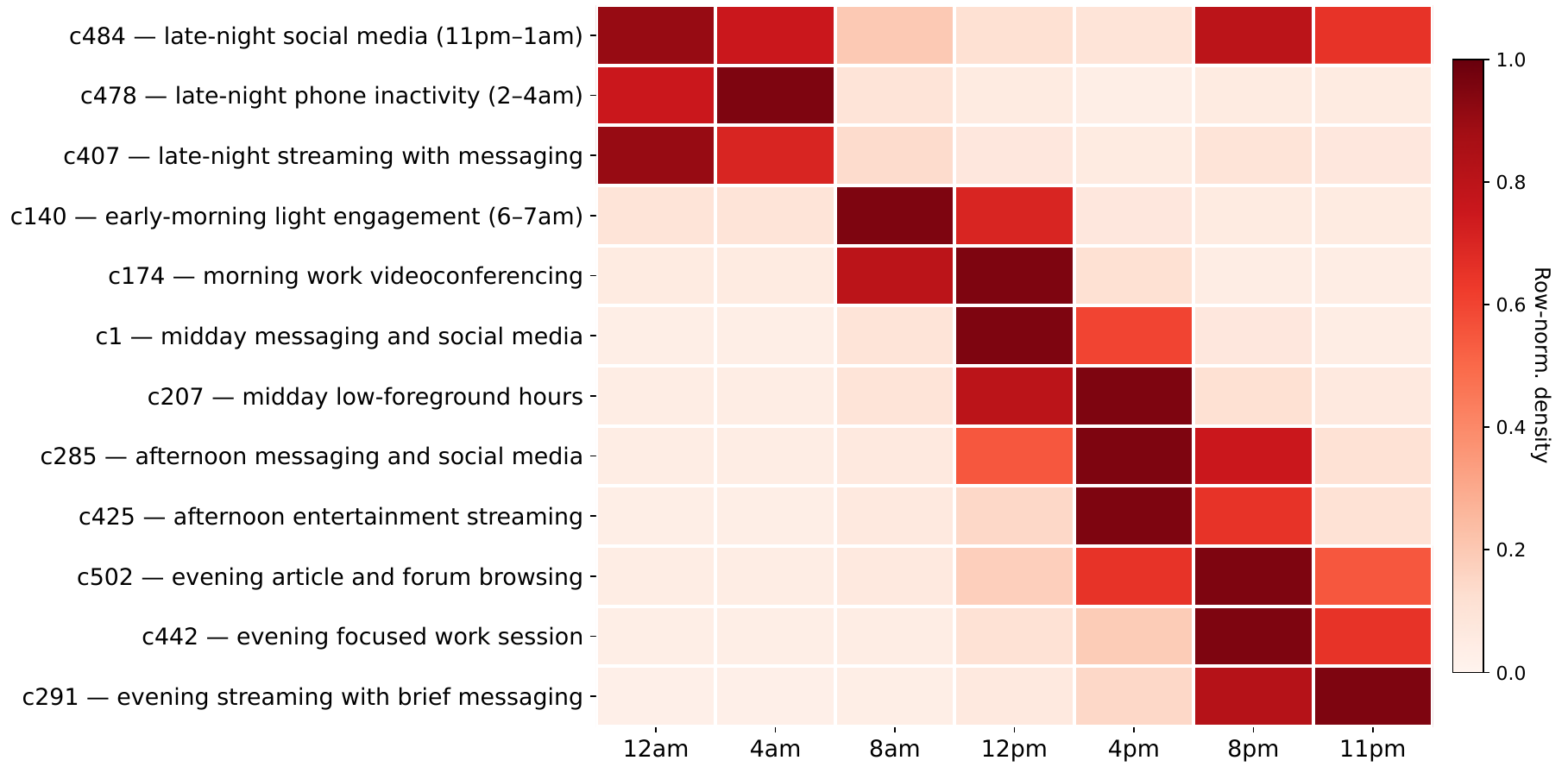}
    \Description{Heatmap with one row per SAE feature and one column per hour of the day. Shading shows row-normalized firing density, and rows are ordered by peak activation hour.}
    \caption{Temporal activation patterns of the 12 SAE features underlying the 14 \textit{robust} associations. Each row shows a feature's row-normalized firing density across the 24-hour day; rows are ordered by peak activation hour, revealing diurnal structure learned without explicit temporal supervision.}
    \label{fig:heatmap}
\end{figure}

Figure~\ref{fig:heatmap} shows the temporal activation patterns of the 12 retained features. Grouping features by their peak activation hour reveals six regimes: night (c484, c478, c407), early morning (c140), morning (c174), midday (c1, c207), afternoon (c285, c425), and evening (c442, c502, c291). The emergence of these distinct diurnal regimes without explicit supervision demonstrates that the SAE recovers recurring behavioral routines directly from the data rather than arbitrary statistical patterns.

To interpret these learned behavioral routines, we examine the dominant foreground applications within each feature's top-100 activating windows (Sec.~\ref{sec:user-diversity-filter}). Table~\ref{tab:feature_profiles} summarizes the resulting behavioral profiles, listing each feature's dominant foreground applications, foreground traffic share, and assigned behavioral label. Features with very low foreground share (e.g., c478 and c207) primarily reflect periods of minimal foreground phone activity rather than a specific application class.

\begin{table}[t]

\caption{Behavioral profiles of the 12 robust SAE features. Foreground share denotes the proportion of bytes in the top-100 activating windows attributable to mapped foreground applications; the remaining traffic corresponds primarily to background system activity. Dominant applications are ordered by foreground traffic share.}

\label{tab:feature_profiles}

\small

\centering

\begin{tabular}{llp{5.4cm}p{4.0cm}}

\toprule

Feature & Foreground share & Dominant foreground applications & Behavioral label \\

\midrule

c1   & 60\% & WeChat, Xiaohongshu, Instagram, Bilibili & Active midday messaging and social media \\

c174 & 91\% & Zoom (80\%), Instagram, Feishu, WeChat & Morning work videoconferencing \\

c478 & 0.6\% & Background OS/CDN dominant & Late-night phone inactivity (2--4am) \\

c285 & 61\% & WeChat (49\%), Xiaohongshu, Netflix, Reddit & Afternoon messaging and social media \\

c484 & 64\% & Tencent, TikTok, Xiaohongshu, Weibo & Late-night social media (11pm--1am) \\

c207 & 5\% & YouTube, Zoom, Instagram & Midday low-foreground activity \\

c291 & 27\% & Netflix, Zoom, WhatsApp, TikTok & Evening streaming with brief messaging \\

c140 & 21\% & Feishu, Instagram, WeChat, YouTube & Early-morning light app engagement (6--7am) \\

c425 & 80\% & Crunchyroll (56\%), YouTube, Instagram & Afternoon entertainment streaming \\

c442 & 79\% & Zoom (55\%), Microsoft (25\%), Adobe & Evening focused work session \\

c502 & 28\% & Reddit, Medium, Spotify, YouTube & Evening article and forum browsing \\

c407 & 92\% & Bilibili, Amazon Video, TikTok, Instagram & Late-night entertainment streaming \\

\bottomrule

\end{tabular}

\end{table}

We validate these interpretations using the perturbation analysis described in Sec.~\ref{sec:user-diversity-filter}. Increasing a feature's activation by $5\times$ its empirical standard deviation consistently shifts the reconstructed behavioral category fractions in the labeled direction, supporting the assigned interpretations. The complete foreground application composition for every feature is provided in Appendix Table~\ref{tab:forensic_labels}.

The behavioral routines described above characterize the learned representation itself rather than their relationship to well-being. To determine which of these learned behavioral routines are associated with longitudinal well-being, we subsequently evaluate each retained feature for associations with sleep disturbance, stress, and loneliness. Only feature--outcome pairs satisfying the predefined statistical significance and robustness criteria are carried forward as the primary findings.

\begin{figure}[htbp]
    \centering
    \includegraphics[width=0.8\textwidth]{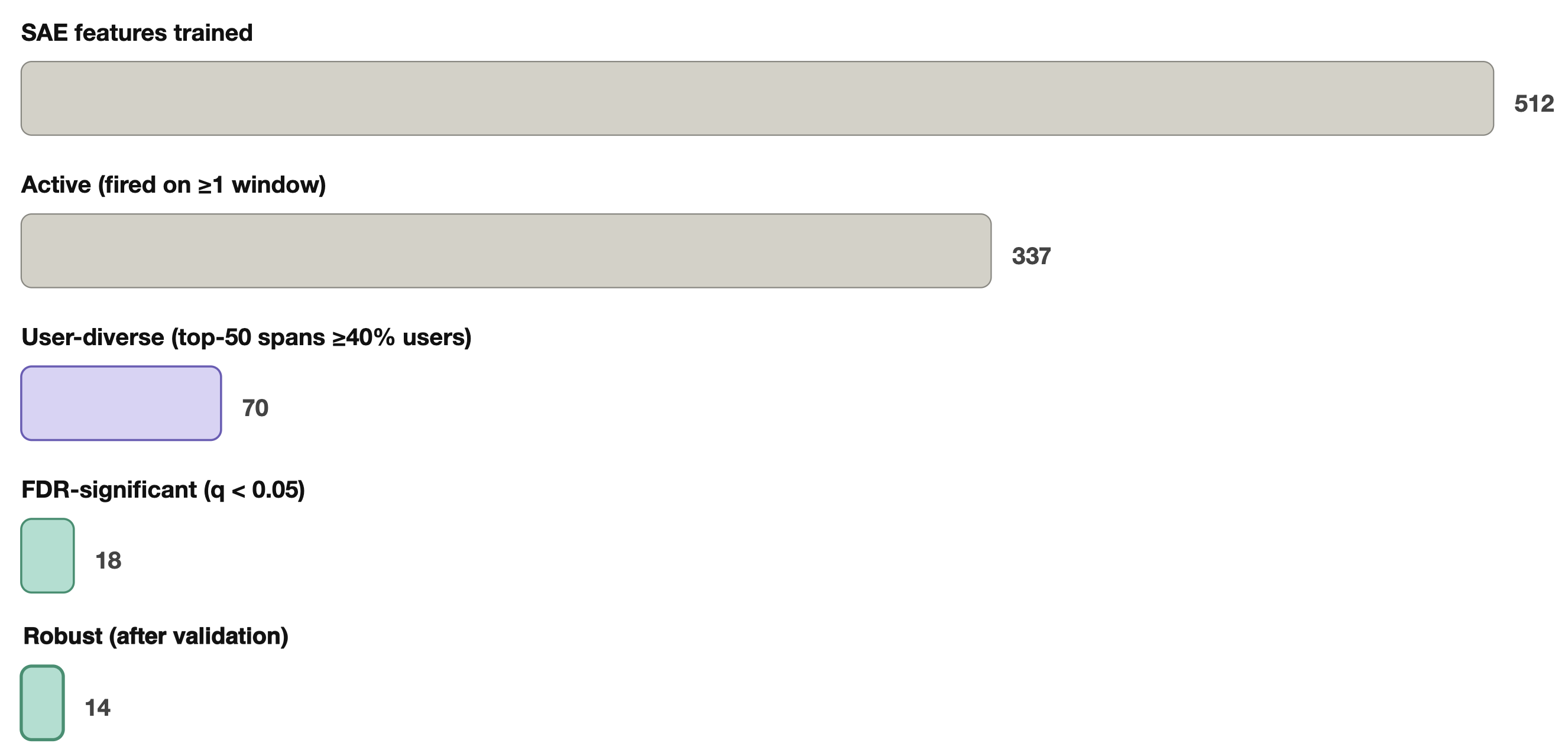}
    \Description{Funnel diagram of feature selection: 512 learned features, 337 active, 70 passing the cross-user generality filter, 18 feature-outcome pairs significant after FDR correction, and 14 robust pairs covering 12 unique features.}
    \caption{SAE feature selection pipeline. From 512 learned features, 337 are active; 70 pass the cross-user generality filter; 18 (feature, outcome) pairs reach FDR significance ($q<0.05$); and 14 pairs (12 unique features) are retained as \textit{Robust} after validation.}
    \label{fig:funnel}
\end{figure}

Figure~\ref{fig:funnel} summarizes this transition from representation learning to longitudinal well-being analysis. Of the 70 candidate features, 18 (feature, outcome) pairs are statistically significant ($q < 0.05$; Sec.~\ref{sec:validation}), and 14 \textit{Robust} associations (spanning 12 unique features) remain after validation (Figure~\ref{fig:outcomes}). These retained feature--outcome associations form the basis of the primary findings presented in Sec.~\ref{sec:main-findings}; the remaining pairs are classified as \textit{Unstable} (Appendix Table~\ref{tab:ablation_full}).

\subsection{What Predefined Network-Traffic Features Capture}
\label{sec:predefined-nt-features}
In Sec.~\ref{sec:recovery-behavioral-metrics-transformer-latents}, we showed that the transformer latents encode known behavioral signals. Here, we ask the converse: how much of the relationship between behavior and well-being can be recovered using these same metrics computed directly from raw network traffic, without any learned representation? These predefined features are computed from the same category-mapped hourly feature representation used as input to the model (Sec.~\ref{sec:feature-construction}). This provides a baseline for assessing the added value of the SAE features, which we evaluate using the same GEE framework (Sec.~\ref{sec:gee}) with FDR correction applied separately to $\beta_B$ and $\beta_W$ (Sec.~\ref{sec:validation}). Results are shown in Table~\ref{tab:classical}.

Five feature–outcome associations survive FDR correction, all at the between-person level ($\beta_B$). Relative amplitude (RA) exhibits strong effects: higher RA—indicating clearer separation between daytime and nighttime activity—is associated with better sleep ($\beta_B = -1.87$, $q = 0.018$) and lower stress ($\beta_B = -1.79$, $q = 0.018$). 

Among behavioral categories, greater communication is associated with lower loneliness ($\beta_B = -0.36$, $q = 0.006$), greater streaming with higher loneliness ($\beta_B = +0.48$, $q = 0.017$), and greater productivity with lower stress ($\beta_B = -0.80$, $q = 0.006$).

\begin{table}[htbp!]
\centering
\caption{Classical behavioral predictors that survived FDR correction ($q < 0.05$) via GEE--Mundlak. All five significant pairs are between-person effects; no within-person classical-feature effect survived FDR. $\beta_B$ = between-person coefficient (standardized). App-category fractions (e.g., communication, streaming) are coarse measures based on destination domains and reflect aggregate activity (e.g., messaging and calling for communication), without distinguishing specific interaction types.}
\label{tab:classical}
\footnotesize
\begin{tabular}{llrc}
\toprule
\textbf{Family} & \textbf{Predictor} & $\beta_B$ & $q$ \\
\midrule
Circadian & Relative amplitude (RA) $\rightarrow$ sleep & $-1.87$ & $0.018$ \\
Circadian & Relative amplitude (RA) $\rightarrow$ stress & $-1.79$ & $0.018$ \\
App fractions & Communication $\rightarrow$ loneliness & $-0.36$ & $0.006$ \\
App fractions & Streaming $\rightarrow$ loneliness & $+0.48$ & $0.017$ \\
App fractions & Productivity $\rightarrow$ stress & $-0.80$ & $0.006$ \\
\bottomrule
\end{tabular}
\end{table}

RA remains significant even though it is not recoverable from the learned representation (Table~\ref{tab:probes}). This is expected: RA depends on rare extrema (the most- and least-active windows), which are smoothed by the transformer’s reconstruction objective, whereas the classical computation preserves them directly from the input time series.

The key finding is what is \emph{not} captured: all five associations are between-person ($\beta_B$) only; no feature shows a significant within-person effect ($\beta_W$). Thus, behavioral summaries computed directly from network traffic distinguish individuals with consistently better or worse outcomes, but fail to capture week-to-week changes within the same individual. In contrast, the SAE-derived features (Sec.~\ref{sec:main-findings}) recover significant within-person associations, demonstrating that the learned representations capture temporal behavioral structure beyond what is available from predefined traffic summaries. This contrast highlights the value of learned representations for longitudinal behavioral sensing.

\subsection{SAE-Derived Behavioral Features and Outcomes}
\label{sec:main-findings}

\begin{figure}[htbp!]
    \centering
    \footnotesize
    \footnotesize
\begin{tabular}{llccccccc}
\toprule
& & \multicolumn{2}{c}{\textbf{Sleep}} & \multicolumn{2}{c}{\textbf{Stress}} & \multicolumn{2}{c}{\textbf{Loneliness}} & \\
\cmidrule(lr){3-4} \cmidrule(lr){5-6} \cmidrule(lr){7-8}
\textbf{Feature} & \textbf{Behavioral pattern} & $\beta_B$ & $\beta_W$ & $\beta_B$ & $\beta_W$ & $\beta_B$ & $\beta_W$ & $\Delta$\textbf{RMSE} \\
\midrule
c1   & active midday messaging and social media          & $+1.94$ & $+0.64$ & $+1.77$ & $+0.46$ & ---     & ---     & 10.1\% / 11.4\% \\
c174 & morning work videoconferencing                    & $-0.65$ & ---     & $-0.43$ & ---     & ---     & ---     & 2.0\% / 1.5\%   \\
c478 & late-night phone inactivity (2--4am)              & ---     & ---     & ---     & ---     & $+0.29$ & $+0.38$ & 1.8\%           \\
\midrule
c285 & afternoon messaging and social media              & ---     & ---     & $+1.29$ & ---     & ---     & ---     & 6.1\%           \\
c484 & late-night social media engagement (11pm--1am)    & ---     & ---     & $+1.61$ & ---     & ---     & ---     & 3.8\%           \\
c207 & midday low-foreground hours                       & ---     & ---     & $-0.84$ & ---     & ---     & ---     & 3.3\%           \\
c291 & evening streaming with brief messaging            & ---     & ---     & $-0.79$ & ---     & ---     & ---     & 4.9\%           \\
c140 & early-morning light foreground engagement (6--7am) & ---    & ---     & $-0.60$ & ---     & ---     & ---     & 3.9\%           \\
c425 & afternoon entertainment streaming                 & ---     & ---     & ---     & ---     & ---     & $-0.32$ & 4.2\%           \\
c442 & evening focused work session                      & ---     & ---     & ---     & ---     & ---     & $-0.22$ & 1.0\%           \\
c502 & evening article and forum browsing                & ---     & ---     & ---     & ---     & ---     & $-0.23$ & 1.7\%           \\
c407 & late-night entertainment streaming with messaging & ---     & ---     & ---     & ---     & ---     & $-0.24$ & 0.9\%           \\
\bottomrule
\end{tabular}
    \includegraphics[width=\textwidth]{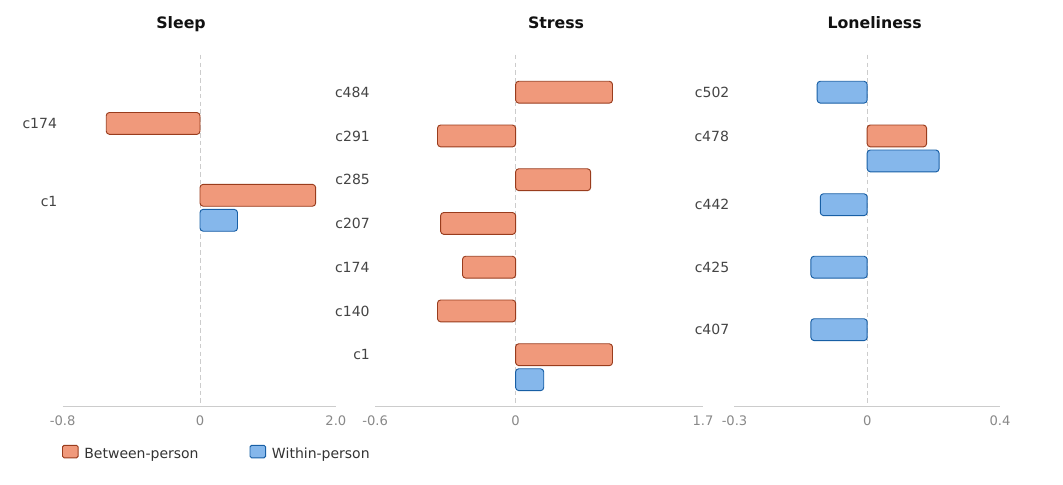}
    \Description{Two panels. The top panel lists the robust SAE feature-outcome associations with their change in RMSE per outcome. The bottom panel plots standardized between-person coefficients in orange and within-person coefficients in blue, grouped by outcome.}
    \caption{(Top) Robust SAE feature–outcome associations. $\Delta$RMSE values are reported per outcome; for features with significant effects on multiple outcomes (c1, c174), both values are shown in column order (sleep / stress). (Bottom) Standardized between- (orange) and within-person (blue) coefficients by outcome domain. Features are ordered by effect magnitude within each panel. Stress is dominated by between-person effects; loneliness by within-person; sleep disturbance shows both.}    \label{fig:outcomes}
\end{figure}

The learned behavioral features characterized in Sec. ~\ref{sec:sae-dervied-behavioral-features} are next evaluated for their associations with sleep disturbance, stress, and loneliness using the GEE--Mundlak framework. This analysis evaluates which learned behavioral patterns are significantly associated with each outcome while separating stable between-person differences from within-person behavioral change. The purpose of this analysis is to characterize longitudinal behavioral associations rather than forecast future well-being trajectories. The following results are restricted to feature--outcome pairs satisfying both statistical significance ($q < 0.05$) and the predefined robustness criteria (Sec.~\ref{sec:validation}). Figure~\ref{fig:outcomes} summarizes these robust associations (top) and visualizes their standardized between- and within-person effects across outcomes (bottom). Behavioral labels are grounded in foreground app composition; detailed per-feature breakdowns are provided in Appendix Table~\ref{tab:forensic_labels}.

Two patterns emerge. First, the dominant type of association differs across outcomes: stress is primarily associated with between-person differences, loneliness with within-person variation, and sleep disturbance reflects a mixture of both. Second, most features exhibit a single type of effect—either between-person or within-person—while only a small subset capture both. Specifically, only three feature–outcome pairs are significant in both components: feature \textit{c1} for sleep disturbance and stress, and feature \textit{c478} for loneliness. These features capture behavioral patterns that both differentiate individuals and track within-person changes over time.
These associations are correlational; the observational design does not permit causal inference, and relationships may reflect bidirectional effects or unobserved confounding.
We next examine each outcome in detail to characterize the specific behavioral patterns underlying these associations.
\subsubsection{Sleep disturbance}
\label{sec:sleep-results}

\begin{figure}[htbp!]
    \centering
    \includegraphics[width=\textwidth]{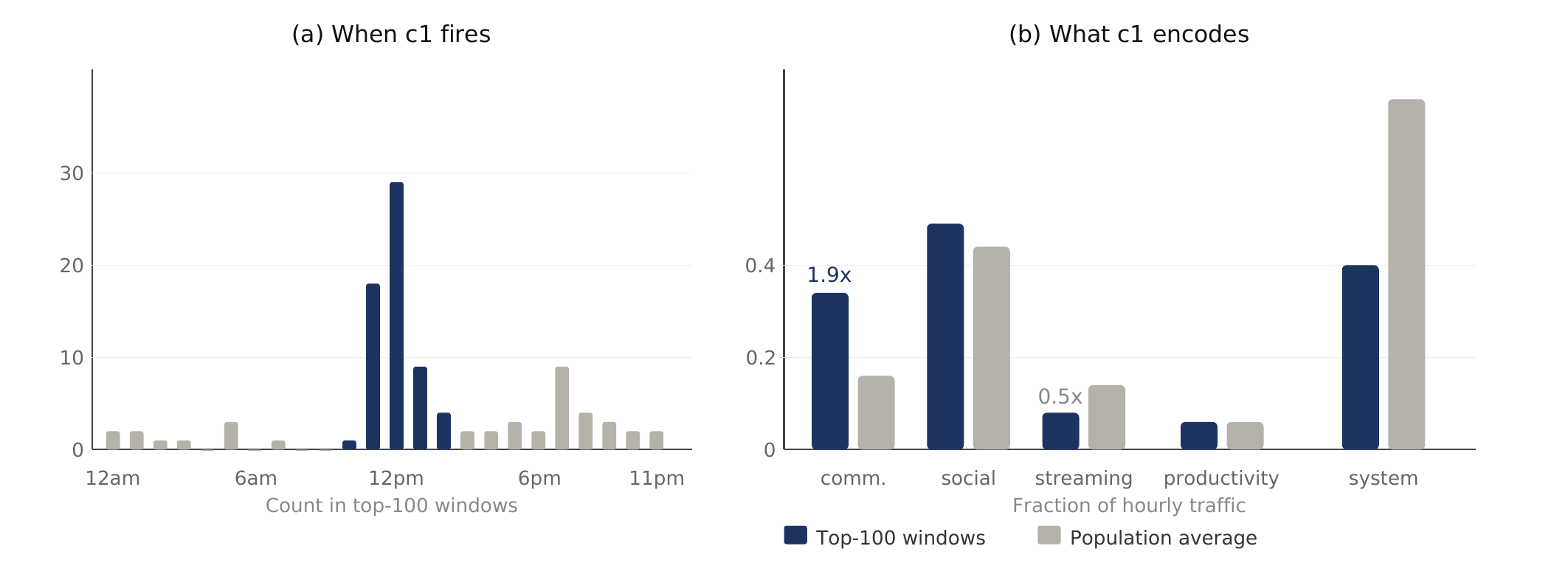}
    \Description{Two panels for feature c1. Panel a shows activation by hour of day with a sharp midday peak. Panel b shows app-category composition relative to the population baseline, with communication elevated and streaming suppressed.}
    \caption{Feature c1 behavioral profile, capturing midday messaging and social media activity. (a) Hour-of-day activation distribution, showing a sharp midday peak. (b) App-category composition relative to population baseline, with elevated communication and suppressed streaming activity.}
    \label{fig:c1}
\end{figure}

Sleep disturbance is associated with two SAE-derived behavioral features that reflect distinct patterns of daily activity.
Feature \textit{c174}, corresponding to morning work videoconferencing (primarily Zoom), exhibits a negative between-person effect ($\beta_B = -0.65$), indicating that individuals who spend a greater proportion of their morning hours in this activity report lower sleep disturbance on average. No within-person effect is observed, suggesting that this association reflects stable differences across individuals rather than week-to-week variation.
In contrast, \textit{c1} captures a midday pattern (10am–2pm) characterized by elevated messaging and social media activity ($1.9\times$ the population mean for communication-category traffic; $0.5\times$ for streaming) (Figure \ref{fig:c1}). Although this pattern occurs infrequently overall (0.4\% of hours), it is observed across 17 of 25 participants, indicating that it is broadly shared.

This feature shows a strong positive association with sleep disturbance at both the between- and within-person levels ($\beta_B = +1.94$, $\beta_W = +0.64$, $q < 0.001$). Individuals with higher overall levels of this midday activity report greater sleep disturbance, and increases relative to an individual’s baseline are likewise associated with worse sleep.

\subsubsection{Stress}
\label{sec:stress-results}

Stress is associated with seven SAE-derived features.  
Higher stress is linked to messaging and social media activity across the day, including \textit{c1} (midday), \textit{c285} (afternoon), and \textit{c484} (late-night, 11pm--1am). Among these, \textit{c1} shows the strongest association and is the only feature with both between- and within-person effects ($\beta_B = +1.77$, $\beta_W = +0.46$): participants with higher overall levels of midday messaging and social media report higher stress, and weeks with increased midday activity relative to baseline are also associated with higher stress. In contrast, \textit{c285} and \textit{c484} exhibit between-person effects only ($\beta_B = +1.29$ and $+1.61$): individuals with more frequent afternoon or late-night social media use report higher stress on average, but week-to-week variation in these patterns does not correspond to changes in stress.  

Lower stress is associated with four features—\textit{c174} (morning videoconferencing), \textit{c207} (midday periods with little active phone use), \textit{c291} (evening streaming with brief messaging), and \textit{c140} (light early-morning phone use, 6--7am)—all of which show negative between-person effects. Individuals whose routines more often include these patterns report lower stress on average, without corresponding within-person associations.

\subsubsection{Loneliness}
Loneliness is primarily associated with within-person variation: week-to-week changes in behavior track changes in reported loneliness. This aligns with prior work characterizing loneliness as a fluctuating state shaped by changing social context \cite{hawkley2010loneliness}.
Four SAE-derived features—\textit{c425} (afternoon entertainment streaming), \textit{c442} (evening focused work sessions), \textit{c502} (evening article and forum browsing), and \textit{c407} (late-night entertainment streaming with messaging)—show negative within-person associations ($\beta_W = -0.22$ to $-0.32$): weeks with higher activation of these patterns are associated with lower loneliness. These features span the afternoon, evening, and late night and correspond to periods of active foreground app use.
In contrast, \textit{c478} shows a positive within-person effect ($\beta_W = +0.38$): weeks with more late-night hours (2--4am) characterized by minimal foreground activity are associated with higher loneliness. These periods are dominated by background system traffic (Appendix Table ~\ref{tab:forensic_labels}), indicating little to no active phone use. Whether this reflects sleep, wakeful disengagement, or both cannot be determined from network traffic alone.
\section{Discussion}
\label{sec:discussion}

Encrypted network traffic is continuously available and works across platforms, and our results show it can support longitudinal behavioral sensing without requiring instrumentation on the device itself. The remainder of this section clarifies the scope, limitations, and implications for future work.

These findings suggest that representation learning enables the recovery of finer-grained behavioral structure from encrypted network traffic than is possible with conventional hand-crafted summaries. Rather than relying on predefined traffic statistics such as session timing, aggregate counts, or application fractions, the transformer and sparse autoencoder discover recurring behavioral routines directly from the traffic. The comparison with the classical baseline demonstrates that these learned representations retain behaviorally meaningful temporal information that supports within-person longitudinal analysis, whereas predefined summaries recover only stable between-person differences. Comparing alternative representation-learning approaches on this signal, such as non-sparse autoencoders or other sequence encoders, is a natural direction for future work. The training objective also determines which behavioral structure is preserved. In longitudinal sensing data, contrastive objectives that draw positive pairs across a person's full record shift representations toward stable identity and away from changing behavioral state \cite{mahmood2026longerrecordsbroaderinvariance}. Masked reconstruction over a bounded 48-hour context is not subject to this particular mechanism, consistent with the within-person associations recovered in Sec. \ref{sec:main-findings}. Accordingly, our objective is to characterize longitudinal behavioral associations rather than forecast future well-being trajectories. This distinction has important implications for digital phenotyping, where many applications—including well-being monitoring—depend on detecting deviations from an individual's own behavioral baseline rather than differences between individuals.

Our modeling approach assumes broadly similar behavioral routines across participants (e.g., nighttime sleep, daytime activity) despite differences in applications or traffic volume. While we show that a single shared backbone generalizes across users with distinct app ecosystems (e.g., WhatsApp/Instagram versus WeChat/Bilibili), this result is established within a relatively homogeneous cohort of university students and may not hold in populations with divergent routines such as shift workers or older adults. Future work should evaluate generalization in such settings and explore approaches that capture subgroup-specific patterns when a single shared backbone is insufficient.

Another limitation lies in our behavioral representation. Activity is expressed as fractions of total traffic volume (Sec.~\ref{sec:feature-construction}), emphasizing high-bandwidth activities (e.g., streaming) while downweighting frequent but low-volume interactions (e.g., browsing or brief app checks). As a result, certain forms of engagement are systematically underrepresented. Incorporating complementary time-based measures—such as session frequency, session duration, or the proportion of active windows—would yield a more balanced characterization of behavior.

Interpretability is further bounded by what network traffic can resolve. SAE features identify which platforms are used and when, but not what users are doing within them—scrolling, posting, or passively consuming content on the same app may produce similar traffic patterns. Consequently, feature interpretations remain indirect, grounded in dominant foreground applications and time-of-day patterns rather than direct observation of user behavior. This limitation is compounded by two factors: user-driven activity cannot be cleanly distinguished from background traffic, and our application-to-category labeling scheme is inherently subjective. Alternative groupings (e.g., separating short-form video from long-form streaming, or distinguishing work from study applications) could yield different interpretations. Scalability is also limited by the manually curated hostname-to-application and application-to-category mappings that underpin our behavioral representation. In this study, these mappings cover 231 hostnames across 92 applications. Although the mappings are reusable across participants and relatively stable over time, they require periodic maintenance as new applications and services emerge, and would need to be extended for deployments spanning different regions or demographic groups with distinct app ecosystems. Future work should integrate network traffic with self-report or externally labeled behavioral data to enable more precise attribution.

A practical limitation of our deployment is its reliance on continuous VPN connectivity. Although modern mobile operating systems support always-on VPN configurations, VPN tunnels may still require manual reactivation following device reboots or operating-system interruptions. This limitation is reflected in our exclusion criteria: 9 of the 42 enrolled participants were excluded because insufficient contiguous data coverage prevented reliable longitudinal analysis.

Finally, the scale and duration of the study constrain the strength of our conclusions. The deployment spans 7 weeks and 25 participants, and should therefore be viewed as hypothesis-generating rather than as providing population-level estimates. Feature activations are often concentrated within a small subset of participants, and although leave-one-out analyses indicate that no single participant drives the observed associations, broader generalization must be established in larger and more diverse cohorts, including older adults, clinical populations, and populations with distinct behavioral routines. A related limitation arises in the latent probing analysis: summarizing representations using means and standard deviations cannot capture behaviors defined by rare extrema (e.g., relative amplitude). This aligns with our observation that such metrics are recoverable from raw traffic but not from the learned representations. Future work should explore probing methods that retain sensitivity to extreme values.

Beyond these technical limitations, our findings also raise broader ethical considerations. Our results demonstrate that encrypted network metadata alone—without access to communication content—can reveal behavioral patterns associated with sleep, stress, and loneliness at the individual level. Importantly, the network vantage point required by our method already exists in practice: Internet Service Providers (ISPs), enterprise networks, and campus networks routinely observe the flow-level metadata used in this work as part of normal network operation. Although we are not aware of evidence that such organizations currently perform behavioral or health inference on these data, these findings suggest that operational network metadata could, in principle, support longitudinal behavioral inference. For example, an organization already observing operational network traffic could apply a pipeline like ours to derive estimates of sleep disturbance, stress, or loneliness without individuals' knowledge or explicit consent. This gap between what users understand they are consenting to and what can actually be inferred from encrypted traffic represents an important ethical risk.

Our study therefore highlights that informed consent alone is insufficient for deployments of network traffic-based behavioral sensing. Users should be informed not only about what network data are collected, but also about the behavioral and health-related inferences that may be derived from those data. In addition, because longitudinal traffic traces and inferred behavioral representations may both be sensitive, deployments should incorporate safeguards such as separating real-world identities from behavioral data, appropriate governance and oversight, and user control over how inferred information is used.

More broadly, these results suggest that learned behavioral representations could serve as a general feature layer for future longitudinal digital phenotyping systems, allowing downstream models to assess a range of behavioral and health outcomes without requiring task-specific hand-engineered features.

\paragraph{Future deployment considerations.}
The present work establishes a proof of concept for longitudinal behavioral sensing from encrypted network traffic. Beyond the methodological contributions presented here, the proposed architecture naturally supports a distributed deployment strategy that may improve scalability and privacy. The shared transformer backbone, which captures population-level behavioral structure, could be hosted centrally, while each user's lightweight adapter could remain on the user's device. Under such a design, personalization would occur locally through the adapter, reducing the need to transmit user-specific behavioral representations or model updates back to a central server. New users could be onboarded by learning only a small adapter while reusing the shared backbone, enabling efficient personalization without retraining the global model. As the user population and application ecosystem evolve, the shared backbone and behavioral mappings could be updated periodically while preserving user-specific adaptation. Organizational deployment would then build upon the governance, ethical, and validation considerations discussed above.
\section{Conclusion}

Smartphones generate continuous streams of encrypted network traffic that have largely been viewed as a source of static behavioral or device-level information. This work demonstrates a broader role for this modality: when analyzed with explicit longitudinal structure and interpretable representation learning, encrypted network metadata can reveal meaningful within-person behavioral dynamics without requiring application instrumentation, operating-system access, or communication content. Rather than simply predicting behavioral states, the resulting representations expose the behavioral concepts underlying those associations and distinguish stable individual differences from changes occurring within the same person over time. The observation that these within-person dynamics cannot be recovered from predefined network-traffic features further suggests that learned representations capture behavioral structure unavailable to hand-engineered approaches. More broadly, these findings reposition encrypted network traffic from a passive networking artifact to a scalable behavioral sensing modality for longitudinal digital health. Although demonstrated here in a university cohort, the framework establishes a foundation for future systems that monitor behavioral change across larger populations and clinical settings while preserving the deployment simplicity of passive network sensing.
\section{Ethical Considerations}
\label{sec:ethics}
This study was approved by the New York University Institutional Review Board (IRB-FY2025-9589). Participants provided informed consent and could withdraw at any time without penalty; participation was offered as an extra-credit option. Data collection was limited to encrypted network metadata and survey responses, with no content or demographic information collected. Data were pseudonymized, stored on access-controlled servers, de-identified prior to analysis, and reported in aggregate.

\begin{acks}
This paper is partly supported by NSF Awards 2219867, 2232655, and 2346332.
\end{acks}

\bibliographystyle{ACM-Reference-Format}
\bibliography{bibliography}
\newpage
\appendix

\renewcommand{\thetable}{A.\arabic{table}}
\renewcommand{\thefigure}{A.\arabic{figure}}
\setcounter{table}{0}
\setcounter{figure}{0}

\section{Participant Inclusion}

\subsection{Participant Flow (CONSORT-style diagram)}
\begin{figure}[htbp]
    \centering
    \includegraphics[width=0.7\columnwidth]{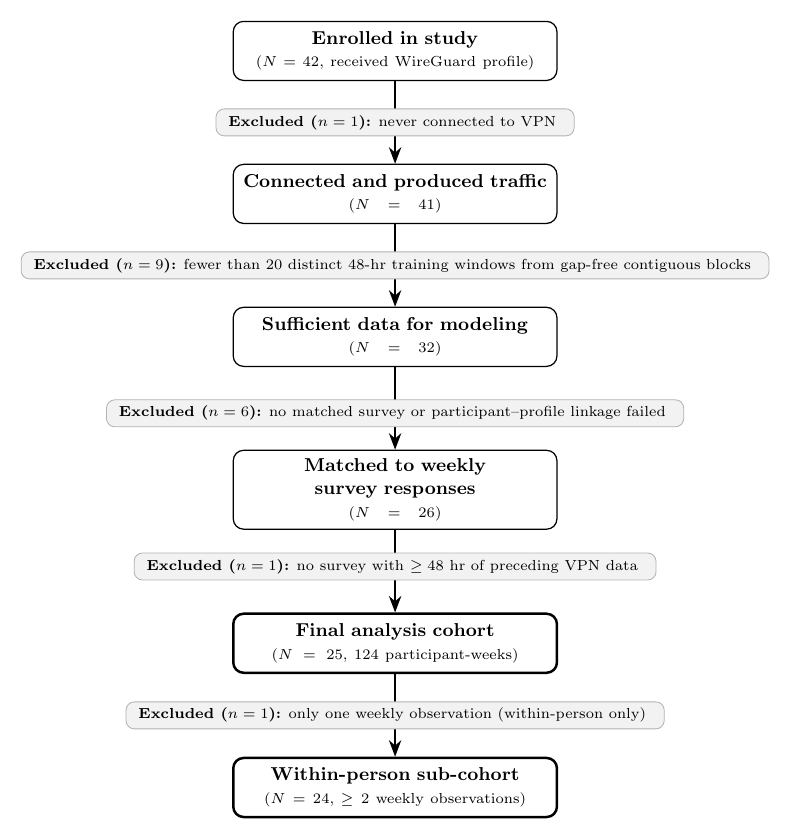}
    \Description{Flow diagram of cohort construction, from 42 enrolled students to 25 in the between-person analysis cohort and 24 in the within-person sub-cohort, with exclusions and reasons at each filtering stage.}
    \caption{Cohort construction. Of 42 students enrolled in the study, 25 contributed to the final between-person 
    analysis cohort and 24 to the within-person sub-cohort. Exclusions 
    and reasons are reported at each filtering stage.}
    \label{app:consort}
\end{figure}

\begin{table}[t]
\centering
\caption{Post hoc psychometric assessment of the self-report outcome measures. Cronbach's $\alpha$ measures the internal consistency of questionnaire items, while the corrected item--total correlation reports the lowest correlation between an individual item and the remaining items within the same scale.}
\label{tab:survey-reliability}
\begin{tabular}{lcccc}
\toprule
\textbf{Outcome} & \textbf{Instrument} & \textbf{Items} & \textbf{Cronbach's $\alpha$} & \textbf{Minimum corrected item--total correlation} \\
\midrule
Sleep disturbance & PROMIS-SD & 4 & 0.69 & 0.40 \\
Perceived stress & PSS-4 & 4 & 0.54 & 0.30 \\
Loneliness & UCLA-3 & 3 & 0.71 & 0.47 \\
\bottomrule
\end{tabular}
\end{table}

\subsection{Window Construction and Coverage Criterion}
\label{sec:week-filter}
Our model (Sec.~\ref{sec:model-architecture}) operates on fixed-length 48-hour sequences of hourly behavioral features, constructed by sliding a window with a 1-hour stride over each participant’s timeline. This requires sufficiently long contiguous periods of reliable data collection. We therefore define an hour as \emph{valid} if at least one network flow is observed. As smartphones generate background traffic even when not actively used (e.g., system services and synchronization), the presence of any flow indicates that the VPN was active; an hour with no observed traffic likely reflects a loss of VPN coverage rather than true inactivity. We partition each participant’s timeline into contiguous blocks of valid hours. A block of length $L$ yields $\max(0, L - 48)$ distinct 48-hour windows under a 1-hour sliding stride. To ensure sufficient data for sequence modeling, we retain participants with at least 20 such windows in total. This ensures that each participant contributes enough data for stable per-user adaptation (Sec.~\ref{sec:model-architecture}). Participants who do not meet this criterion are excluded due to insufficient contiguous coverage.

\section{Dictionary Construction and Category Design}
\label{app:dictionary}

\subsection{Dictionary Construction}

The hostname-to-application dictionary was constructed through a 
consensus-based process. One author created the initial hostname-to-
application and application-to-category mappings, and a second author 
independently reviewed them for consistency.

Applications may be associated with multiple domains (e.g., 
\texttt{whatsapp.com}, \texttt{whatsapp.net}). In addition, some domains 
host multiple services across subdomains (e.g., 
\texttt{mail.google.com} $\rightarrow$ \textit{gmail}, 
\texttt{drive.google.com} $\rightarrow$ \textit{google\_drive}), which 
are disambiguated at the subdomain level where possible.

\subsection{Category Definition}

We define 11 behavioral categories representing broad types of user 
activity (e.g., communication, social media, streaming, and 
productivity). These categories are designed to reflect functional 
behavior rather than platform-specific taxonomies such as app store 
genres. One of the 11 categories, \textit{other}, is a residual bucket for hostnames that do not map to any named category; it is retained for dictionary completeness but is not analyzed as a behavioral category.

Each application is assigned to a single category based on its primary 
functional use (e.g., interpersonal communication, content consumption, 
or productivity). For example, messaging platforms such as WhatsApp and 
Messenger are categorized as \textit{communication}, while platforms 
such as Facebook and Instagram are categorized as \textit{social media}, 
reflecting their primary role in feed-based interaction.

Although some applications span multiple functions, assigning each 
application to a single category ensures interpretability and avoids 
overlapping feature definitions.

\subsection{Validation Against External Taxonomies}

To validate our application-to-category assignments, we compared them 
against Apple App Store genres using the iTunes Search API 
\cite{apple_itunes_search_api}.

We evaluated all consumer-facing applications in the five model-input 
categories (39 applications), i.e., the subset for which App Store labels 
are defined and interpretable. Applications in excluded categories, 
system services (e.g., iCloud, Google APIs), and background services 
without App Store entries were not included.

A contradiction was defined as a case where an application assigned to 
one behavioral category (e.g., \textit{communication}) was labeled by 
Apple under an unrelated domain (e.g., Finance or Games). No such 
contradictions were observed.

We emphasize that App Store genres are not used as ground truth, as they 
reflect content type rather than behavioral function. Instead, this 
comparison serves as a consistency check to identify potential 
misclassifications.

\subsection{Hostname-to-Application Coverage}

The resulting dictionary accounts for approximately 77\% of total 
observed traffic volume. The remaining 23\% corresponds to a long tail 
of services that are not mapped to any behavioral category. This 
unmapped traffic is not treated as a separate feature; instead, it is 
included only in the total traffic used to normalize category-level 
features.

\section{Category Selection Criteria and Sensitivity}
\label{app:category-selection}

\subsection{Selection Criteria}

Categories are evaluated using two criteria:
(i) \emph{participant coverage}, defined as the fraction of participants 
with non-zero weekly activity in at least half of their observed weeks, 
and (ii) \emph{traffic contribution}, defined as the fraction of total 
bidirectional traffic volume.

We require each category to have at least 40\% participant coverage and 
contribute at least 1\% of total traffic. These thresholds are chosen to 
ensure stable estimation when features are represented as weekly 
fractions of total traffic.

\subsection{Rationale for Exclusion}

Categories with low coverage or low traffic volume produce unstable 
feature estimates at weekly resolution. When a category contributes a 
very small number of bytes, even minor absolute changes in activity can 
lead to large relative fluctuations in its fraction of total traffic. 
Similarly, categories observed in only a small subset of participants 
result in between-person comparisons driven by a few individuals rather 
than reflecting population-level patterns.

\subsection{Excluded Categories}

\textit{Gaming} fails the coverage criterion. Although it accounts for 
1.2\% of total traffic, only 32\% of participants generate any gaming 
traffic, and a single participant contributes 79\% of all gaming bytes.

\textit{Browsing} and \textit{shopping} fail the volume criterion. 
Despite being highly prevalent (95\% and 93\% of participants, 
respectively), they contribute only 0.7\% and 0.4\% of total traffic.

These categories therefore produce unstable weekly estimates and are 
excluded from downstream modeling.

\begin{table}[t]
\centering
\caption{Complete architecture and training hyperparameters for the transformer backbone, per-user adapter, and sparse autoencoder (SAE). Primary SAE configuration is reported; a larger 2048/$k=32$ variant was also evaluated but is not used for primary findings.}
\label{tab:hyperparams}
\small
\begin{tabular}{lll}
\toprule
\textbf{Component} & \textbf{Parameter} & \textbf{Value} \\
\midrule
\multirow{7}{*}{Transformer} & Input dim & 8 \\
 & Embedding dim ($d_{\text{model}}$) & 64 \\
 & Attention heads & 4 \\
 & Transformer layers & 2 \\
 & Positional encoding & sinusoidal \\
 & Context window & 48 hours \\
 & Prediction head & Linear($64 \to 8$), shared \\
\midrule
\multirow{2}{*}{User adapter} & Architecture & Linear$(64,64)$ + ReLU + Linear$(64,64)$ \\
 & Composition & residual: $h' = h + \mathrm{MLP}(h)$ \\
\midrule
\multirow{6}{*}{Training} & Train/test split & chronological, $70/30$ per user \\
 & Optimizer & Adam \\
 & Learning rate & $1 \times 10^{-3}$ \\
 & Batch size & 64 \\
 & Phase 1 (backbone + head) & 30 epochs, all users \\
 & Phase 2 (adapter, per user) & 15 epochs per user \\
\midrule
\multirow{7}{*}{SAE (primary)} & Dictionary size & 512 \\
 & Active features per sample ($k$) & 16 \\
 & Optimizer & Adam \\
 & Learning rate & $3 \times 10^{-4}$ \\
 & LR schedule & cosine annealing \\
 & Batch size & 512 \\
 & Epochs & 500 \\
\midrule
\multirow{3}{*}{Validation} & SAE held-out users & 5 (randomly selected, seed 42) \\
 & Checkpoint selection & lowest validation MSE \\
 & Decoder columns & re-normalized to unit $L_2$ after every step \\
\midrule
\multirow{2}{*}{Reproducibility} & Global seed & 42 \\
 & cuDNN deterministic mode & True \\
\bottomrule
\end{tabular}
\end{table}
\begin{table}[t]
\centering
\caption{Threshold-sensitivity analysis. Number of pairs classified \textsc{Robust} as the $\Delta\mathrm{RMSE}$ and sign-consistency thresholds are varied over a $7 \times 6$ grid. The primary thresholds ($\Delta\mathrm{RMSE} > 0.5\%$, sign $> 60\%$) are shown in bold and yield 14 pairs. Three pairs---c1 for sleep disturbance and stress, and c285 for stress---remain \textsc{Robust} across all threshold combinations, including the strictest setting ($\Delta\mathrm{RMSE} > 5.0\%$, sign $> 75\%$).}
\label{tab:thresh_sweep}
\small
\begin{tabular}{lcccccc}
\toprule
 & \multicolumn{6}{c}{\textbf{Sign consistency}} \\
\cmidrule(lr){2-7}
$\Delta\mathrm{RMSE}$ & $>50\%$ & $>55\%$ & $>60\%$ & $>65\%$ & $>70\%$ & $>75\%$ \\
\midrule
$>0.5\%$ & 18 & 16 & \textbf{14} & 14 & 11 & 9 \\
$>1.0\%$ & 15 & 14 & 12 & 12 & 10 & 8 \\
$>1.5\%$ & 15 & 14 & 12 & 12 & 10 & 8 \\
$>2.0\%$ & 11 & 10 & 9 & 9 & 7 & 6 \\
$>2.5\%$ & 8 & 8 & 8 & 8 & 6 & 6 \\
$>3.0\%$ & 8 & 8 & 8 & 8 & 6 & 6 \\
$>5.0\%$ & 3 & 3 & 3 & 3 & 3 & 3 \\
\bottomrule
\end{tabular}
\end{table}

\begin{table*}[t]
\centering
\caption{\textbf{Sensitivity of the sparse autoencoder decomposition to the TopK parameter $k$.} For each value of $k$, we repeated the complete feature-selection and validation pipeline. The principal findings are stable in a mid-sparsity regime ($k=8$--16). At lower $k$ the decomposition recovers fewer robust behavioral features, whereas higher $k$ produces increasingly fragmented decompositions. The purpose of this analysis is to evaluate sensitivity to the TopK parameter rather than to optimize it; the primary configuration ($k=16$) was fixed before downstream statistical analysis.}
\label{tab:topk_sensitivity}
\centering
\begin{tabular}{rccccp{5cm}}
\toprule
\textbf{$k$} & \textbf{Active SAE} & \textbf{User-general} & \textbf{FDR-significant} & \textbf{Robust} & \textbf{Interpretation} \\
 & \textbf{features} & \textbf{features} & \textbf{pairs} & \textbf{associations} & \\
\midrule
4  & 217 & 53 & 9  & 4  & Dictionary too sparse to isolate behavioral concepts \\
8  & 294 & 82 & 28 & 18 & Stable decomposition \\
\textbf{16} & \textbf{337} & \textbf{70} & \textbf{18} & \textbf{14} & \textbf{Stable decomposition (primary configuration)} \\
32 & 404 & 59 & 8  & 5  & Behavioral concepts fragment across multiple latent features \\
64 & 267 & 54 & 7  & 4  & Behavioral concepts fragment across multiple latent features \\
\bottomrule
\end{tabular}
\end{table*}

\newpage
\section{Label Threshold Sensitivity}
\label{app:threshold-sens}

We assessed the stability of feature labels 
(Sec.~\ref{sec:user-diversity-filter}) by sweeping the high cutoff 
over $\{1.3, 1.4, 1.5, 1.6, 1.7, 1.8, 2.0\}$ and the low cutoff over 
$\{0.60, 0.65, 0.70, 0.75, 0.80\}$, yielding 35 valid threshold pairs. 
For each pair, we re-labeled the $12 \times 5 = 60$ (feature, category) 
cells of the robust feature set and computed the fraction whose label 
matches the canonical $(1.5, 0.7)$ assignment. 
Table~\ref{tab:threshold-sens} reports the full sweep. Across all 35 
pairs, $\geq 83\%$ of cells retained their canonical label (median 
95\%); no cell ever flipped between \emph{high} and \emph{low}, and 
every change was between a labeled state (\emph{high} or \emph{low}) 
and \emph{neutral}. Per-feature stability was also high: 1 of 12 
robust features (c478) showed no label changes across any threshold 
pair, 9 features showed at most one (feature, category) label change 
across the entire sweep, and the remaining 2 features (c285 and c407) 
showed at most two changes out of 5 category labels each.

\begin{table}[htbp!]
\centering
\caption{Label threshold sensitivity sweep. For each pair of high 
and low cutoff multipliers, we re-labeled the $12 \times 5 = 60$ 
(feature, category) cells of the robust feature set and computed 
the fraction whose label matches the canonical $(1.5, 0.7)$ 
assignment (bolded). ``\# changed'' counts cells whose label 
differs from the canonical assignment. No cell ever flipped between 
\emph{high} and \emph{low} under any threshold pair; all changes 
are between a labeled state and \emph{neutral}.}
\label{tab:threshold-sens}
\small
\begin{tabular}{cccc}
\toprule
high & low & \% unchanged & \# changed \\
\midrule
1.3 & 0.60 & 93.3\% & 4 \\
1.3 & 0.65 & 93.3\% & 4 \\
1.3 & 0.70 & 95.0\% & 3 \\
1.3 & 0.75 & 90.0\% & 6 \\
1.3 & 0.80 & 86.7\% & 8 \\
1.4 & 0.60 & 98.3\% & 1 \\
1.4 & 0.65 & 98.3\% & 1 \\
1.4 & 0.70 & 100.0\% & 0 \\
1.4 & 0.75 & 95.0\% & 3 \\
1.4 & 0.80 & 91.7\% & 5 \\
1.5 & 0.60 & 98.3\% & 1 \\
1.5 & 0.65 & 98.3\% & 1 \\
\textbf{1.5} & \textbf{0.70} & \textbf{100.0\%} & \textbf{0} \\
1.5 & 0.75 & 95.0\% & 3 \\
1.5 & 0.80 & 91.7\% & 5 \\
1.6 & 0.60 & 98.3\% & 1 \\
1.6 & 0.65 & 98.3\% & 1 \\
1.6 & 0.70 & 100.0\% & 0 \\
1.6 & 0.75 & 95.0\% & 3 \\
1.6 & 0.80 & 91.7\% & 5 \\
1.7 & 0.60 & 95.0\% & 3 \\
1.7 & 0.65 & 95.0\% & 3 \\
1.7 & 0.70 & 96.7\% & 2 \\
1.7 & 0.75 & 91.7\% & 5 \\
1.7 & 0.80 & 88.3\% & 7 \\
1.8 & 0.60 & 95.0\% & 3 \\
1.8 & 0.65 & 95.0\% & 3 \\
1.8 & 0.70 & 96.7\% & 2 \\
1.8 & 0.75 & 91.7\% & 5 \\
1.8 & 0.80 & 88.3\% & 7 \\
2.0 & 0.60 & 90.0\% & 6 \\
2.0 & 0.65 & 90.0\% & 6 \\
2.0 & 0.70 & 91.7\% & 5 \\
2.0 & 0.75 & 86.7\% & 8 \\
2.0 & 0.80 & 83.3\% & 10 \\
\bottomrule
\end{tabular}
\end{table}

\newpage

\section{Classical Feature Definitions}
\label{app:classical-features}

All classical features are computed from the same hourly activity 
time series $a_h$ ($h = 0, 1, \ldots, H-1$), where $a_h$ denotes 
the per-user normalized flow count (\texttt{flow\_count\_pct}; 
Sec.~\ref{sec:feature-construction}) at hour $h$ and $H$ is the 
number of hours in the observation window. Let $\bar{a}$ denote 
the mean of $a_h$ over the window.

\subsection{Circadian rhythm measures.} Following the rest-activity 
rhythm literature \cite{van1999bright, witting1990alterations, gonccalves2015fresh}:

\subsubsection{Interdaily stability (IS)} IS measures the day-to-day 
consistency of the activity pattern at the same hour of day:
\[
\mathrm{IS} = 
\frac{H \sum_{h=0}^{23} (\bar{a}_h - \bar{a})^2}
{24 \sum_{h=0}^{H-1} (a_h - \bar{a})^2},
\]
where $\bar{a}_h$ is the mean activity at hour-of-day 
$h \in \{0, \ldots, 23\}$ averaged over days. 
$\mathrm{IS} \in [0, 1]$, with higher values indicating more 
stable circadian rhythms.

\subsubsection{Intradaily variability (IV)} IV measures the fragmentation of 
the activity pattern, capturing transitions between high and low 
activity within days:
\[
\mathrm{IV} = 
\frac{H \sum_{h=1}^{H-1} (a_h - a_{h-1})^2}
{(H-1) \sum_{h=0}^{H-1} (a_h - \bar{a})^2}.
\]
Higher values indicate more fragmented rhythms.

\subsubsection{Relative amplitude (RA)} RA measures the contrast between the 
most-active and least-active periods of the observation window:
\[
\mathrm{RA} = \frac{M_{10} - L_5}{M_{10} + L_5},
\]
where $L_5$ is the minimum mean of any 5 consecutive hours and 
$M_{10}$ is the maximum mean of any 10 consecutive hours within 
the observation window (computed as rolling windows over the full 
window, not within individual days). $\mathrm{RA} \in [0, 1]$, 
with higher values indicating greater rest-activity contrast.

\subsection{Time-of-day activity statistics.} 

We partition the day into four contiguous 6-hour bands: 
night ($\{0, \ldots, 5\}$), morning ($\{6, \ldots, 11\}$), 
afternoon ($\{12, \ldots, 17\}$), and evening 
($\{18, \ldots, 23\}$). For each band $B$, we compute the 
mean activity within the band:
\[
\bar{a}_B = \frac{1}{|B|} \sum_{h \in B} a_h.
\]
We additionally compute the \textbf{night-to-morning ratio} 
$\bar{a}_{\text{night}} / \bar{a}_{\text{morning}}$, capturing 
the relative balance of late-night and morning activity.

\subsection{Activity timing.}

\subsubsection{Activity centroid (center of mass)} The activity-weighted 
mean hour-of-day,
\[
\mathrm{COM} = \frac{\sum_{h=0}^{23} h \cdot p_h}{\sum_{h=0}^{23} p_h},
\]
where $p_h$ is the cross-day mean activity at hour-of-day $h$. The 
COM summarizes when, on average, a participant's activity is 
concentrated.

\subsubsection{Active span} The duration (in hours) of the participant's 
typical active period, computed as 
$(\text{offset} - \text{onset}) \bmod 24$, where onset is the first 
hour-of-day and offset is the last hour-of-day at which the 
cross-day-averaged activity profile exceeds the 25th percentile of 
its nonzero values.

\subsection{Weekly structure.}

Weekday--weekend activity difference: 
$\bar{a}_{\text{weekday}} - \bar{a}_{\text{weekend}}$, where 
$\bar{a}_{\text{weekday}}$ is the mean activity over Mon--Fri and 
$\bar{a}_{\text{weekend}}$ is the mean activity over Sat--Sun within 
the observation window.

\section{Validation and Robustness Checks}
\subsection{Multiple-Testing Correction}
\label{sec:fdr}
As we test many feature–outcome pairs, some associations are expected to appear significant purely by chance, increasing the risk of Type I error. To control for this, we apply Benjamini--Hochberg false discovery rate (FDR) correction \cite{benjamini1995controlling} separately to between-person ($\beta_B$) and within-person ($\beta_W$) effects.

Within each effect type, p-values are pooled across all feature–outcome pairs before correction, so that significance is evaluated relative to the total number of tests performed. A feature is considered significant if either its between-person or within-person effect satisfies $q < 0.05$, a standard threshold that controls the expected proportion of false discoveries among the reported significant results.

We apply the correction separately for between-person ($\beta_B$) and within-person ($\beta_W$) effects because they address distinct questions—whether stable differences across individuals are associated with outcomes, and whether deviations from an individual’s baseline are associated with changes in outcomes.

\subsection{Post Hoc Validation of Significant Features}
\label{sec:ablation-falsification-test}

Statistical significance alone does not guarantee that a feature captures a meaningful or reliable behavioral signal. To assess whether FDR-significant features provide substantive predictive information, we evaluate each (feature, outcome) pair against a simple baseline.

For each significant pair, we use the fitted univariate GEE model (described in Sec.~\ref{sec:gee}) and compare it to an intercept-only null model that ignores the feature and predicts the same value (the grand mean $\bar{Y}$) for all observations. Prediction error is measured using root mean squared error (RMSE):
\[
\mathrm{RMSE} = \sqrt{\tfrac{1}{N}\sum_{i,t}(Y_{it} - \hat{Y}_{it})^2}.
\]

The contribution of each feature is quantified as the relative reduction in error:
\[
\Delta\mathrm{RMSE}\% =
100 \cdot
\frac{\mathrm{RMSE}_\text{null} - \mathrm{RMSE}_\text{model}}
{\mathrm{RMSE}_\text{null}}.
\]

To assess whether a feature produces consistent effects, we examine whether the model’s predictions tend to lie consistently above or below the grand mean of the outcome ($\bar{Y}$). Specifically, we compute
\[
\delta_{it} = \hat{Y}_{it} - \bar{Y},
\]
where $\hat{Y}_{it}$ is the model’s prediction and $\bar{Y}$ is the mean of the outcome across all observations. A positive value indicates that the feature shifts predictions above the overall mean, while a negative value indicates a shift below the mean. We then calculate the proportion of observations for which $\delta_{it}$ has the same sign. High consistency indicates that the feature reliably shifts predictions in one direction, rather than producing mixed or unstable effects across observations.

Based on predictive contribution ($\Delta\mathrm{RMSE}\%$) and directional consistency, each (feature, outcome) pair is classified into one of four categories. A feature is labeled \textit{Robust} if it reduces prediction error ($\Delta\mathrm{RMSE}\% > 0.5\%$) and exhibits consistent directional effects (sign consistency $> 60\%$). Features that reduce error but show inconsistent direction are labeled \textit{Unstable}, while those with consistent direction but negligible error reduction are labeled \textit{Redundant}. All remaining features are classified as \textit{Noise}. Only \textit{Robust} features are reported as primary findings; all others are provided in Appendix Table~\ref{tab:ablation_full}.

\subsection{Robustness and Sensitivity Analyses}
\label{sec:robustness-sensitivity-analyses}

We evaluate our findings using two complementary analyses.

First, we perform \textit{leave-one-user-out (LOUO) resampling}. For each FDR-significant (feature, outcome) pair, we refit the GEE model (from Sec. \ref{sec:gee}) while holding out one participant at a time, and examine whether the estimated effect retains the same direction and remains statistically significant across folds. This tests whether results depend disproportionately on any single participant.

Second, we conduct a \textit{threshold sensitivity analysis}. As the classification of Robust features depends on thresholds for $\Delta\mathrm{RMSE}$ and sign consistency, we vary these thresholds over a range of values and recompute the number of features classified as Robust at each setting. This assesses whether our findings are stable to the choice of these thresholds.
\begin{table*}[htbp]
\centering
\caption{Full validation verdicts for all 18 FDR-significant (feature, outcome) pairs. A pair is classified \textsc{Robust} if $\Delta\mathrm{RMSE} > 0.5\%$ and sign consistency $> 60\%$; \textsc{Unstable} if $\Delta\mathrm{RMSE} > 0.5\%$ but sign consistency $\leq 60\%$. No pairs fell into the \textsc{Redundant} or \textsc{Noise} categories at these thresholds. Only \textsc{Robust} verdicts are used as primary findings.}
\label{tab:ablation_full}
\small
\begin{tabular}{llllll}
\toprule
\textbf{Feature} & \textbf{Outcome} & \textbf{Effect} & $\Delta$\textbf{RMSE} & \textbf{Sign} & \textbf{Verdict} \\
\midrule
c407 & Loneliness       & $\beta_W = -0.24$                        & $0.94\%$ & $68\%$ & \textsc{ROBUST} \\
c425 & Loneliness       & $\beta_W = -0.32$                        & $4.23\%$ & $81\%$ & \textsc{ROBUST} \\
c442 & Loneliness       & $\beta_W = -0.22$                        & $0.99\%$ & $94\%$ & \textsc{ROBUST} \\
c478 & Loneliness       & $\beta_B = +0.29$ / $\beta_W = +0.38$    & $1.78\%$ & $79\%$ & \textsc{ROBUST} \\
c502 & Loneliness       & $\beta_W = -0.23$                        & $1.66\%$ & $92\%$ & \textsc{ROBUST} \\
c1 & Sleep disturbance      & $\beta_B = +1.94$ / $\beta_W = +0.64$    & $10.10\%$ & $84\%$ & \textsc{ROBUST} \\
c174 & Sleep disturbance      & $\beta_B = -0.65$                        & $2.01\%$ & $75\%$ & \textsc{ROBUST} \\
c1 & Stress     & $\beta_B = +1.77$ / $\beta_W = +0.46$    & $11.39\%$ & $85\%$ & \textsc{ROBUST} \\
c140 & Stress     & $\beta_B = -0.60$                        & $3.88\%$ & $81\%$ & \textsc{ROBUST} \\
c174 & Stress     & $\beta_B = -0.43$                        & $1.50\%$ & $75\%$ & \textsc{ROBUST} \\
c207 & Stress     & $\beta_B = -0.84$                        & $3.32\%$ & $68\%$ & \textsc{ROBUST} \\
c285 & Stress     & $\beta_B = +1.29$                        & $6.11\%$ & $83\%$ & \textsc{ROBUST} \\
c291 & Stress     & $\beta_B = -0.79$                        & $4.89\%$ & $69\%$ & \textsc{ROBUST} \\
c484 & Stress     & $\beta_B = +1.61$                        & $3.81\%$ & $84\%$ & \textsc{ROBUST} \\
c175 & Loneliness       & $\beta_W = -0.36$                        & $0.86\%$ & $54\%$ & \textsc{UNSTABLE} \\
c286 & Loneliness       & $\beta_W = -0.66$                        & $2.14\%$ & $58\%$ & \textsc{UNSTABLE} \\
c330 & Loneliness       & $\beta_W = -0.57$                        & $2.28\%$ & $52\%$ & \textsc{UNSTABLE} \\
c501 & Stress     & $\beta_W = -0.93$                        & $1.72\%$ & $60\%$ & \textsc{UNSTABLE} \\
\bottomrule
\end{tabular}
\end{table*}

\section{Per-User Adapter Specialization}
\label{app:adapter-specialization}

A key concern is whether participants with atypical behavior could 
contaminate the shared representation, biasing it toward those 
individuals. Our architecture mitigates this through two distinct 
mechanisms targeting different parts of the pipeline. For the 
masked-reconstruction model, per-user residual adapters provide 
additive corrections on top of the shared backbone, allowing the 
backbone to capture population-level temporal structure while 
user-specific deviations are absorbed by each participant's own 
adapter. For the downstream SAE features (which operate on 
backbone-only encodings and do not pass through the adapters), 
the user-diversity filter (Sec.~\ref{sec:user-diversity-filter}) 
removes any feature whose activations are concentrated in fewer 
than 40\% of participants, and the LOUO resampling check 
(Sec.~\ref{sec:validation}) verifies that downstream associations 
do not depend on any single participant. Together, these mechanisms 
guard against contamination at both the representation and feature 
levels.

We empirically verify the adapter specialization first. We examine 
the per-user adapter deltas, 
$\delta_u(x) = \text{adapter}_u(\text{backbone}(x)) - \text{backbone}(x)$, 
which capture the transformation applied for each participant. The 
deltas are nearly orthogonal across users (mean pairwise cosine 
similarity 0.02 across all 496 trainable-user pairs; no pair 
exceeds 0.5; 43\% of pairs are negatively correlated), indicating 
that each adapter encodes a distinct, user-specific adjustment 
rather than introducing shared structure into the backbone.

Second, the leave-one-subject-out linear-probe analysis from 
Sec.~\ref{sec:recovery-behavioral-metrics-transformer-latents} 
provides evidence that the backbone latents generalize across 
participants. In each fold, a linear probe fit on the other 
participants' (latent, metric) pairs recovers the held-out 
participant's behavioral metric from their own latents (e.g., 
morning activity: $r = 0.75$; night-to-morning ratio: $r = 0.74$; 
interdaily stability: $r = 0.47$). The backbone itself was trained 
once on the full cohort; the held-out participant is excluded only 
from the probe-fitting step. This generalization of the 
latent$\to$metric mapping indicates the backbone latents encode 
behaviorally interpretable structure that is shared across 
participants, not idiosyncratic to particular individuals.

\begin{table*}[htbp]
\caption{Behavioral interpretation of robust SAE features. Each row reports the dominant foreground app composition in the feature's top-100 activating windows and the behavioral label used in the main text (Table~\ref{tab:feature_profiles}). Foreground share (\%) is the proportion of bytes attributable to mapped foreground apps; remaining bytes correspond to background system traffic (e.g., OS sync, CDN). Top foreground apps are ordered by share within the foreground portion. Behavioral labels are grounded in this foreground composition; features with low foreground share (e.g., c478, c207) primarily capture absence of human-driven activity, with the labels reflecting the foreground content where present.}
\label{tab:forensic_labels}
\footnotesize
\begin{tabular}{p{0.5cm}p{0.7cm}p{6.5cm}p{6.0cm}}
\toprule
\textbf{Feat.} & \textbf{Fg \%} & \textbf{Top foreground apps} & \textbf{Behavioral label (main text)} \\
\midrule
c1   & 60\%  & WeChat (27\%); Xiaohongshu (20\%); gaming (14\%); Instagram (12\%); Bilibili (4\%)    & active midday messaging and social media \\
c174 & 91\%  & Zoom (80\%); Instagram (5\%); Feishu (3\%); WeChat (2\%); Microsoft (2\%)             & morning work videoconferencing \\
c478 & 0.6\% & Background OS/CDN traffic dominant; foreground near zero across all categories       & late-night phone inactivity (2--4am) \\
\midrule
c285 & 61\%  & WeChat (49\%); Xiaohongshu (21\%); Netflix (5\%); Reddit (3\%); Spotify (3\%)         & afternoon messaging and social media \\
c484 & 64\%  & Tencent (37\%); TikTok (13\%); Xiaohongshu (13\%); Weibo (8\%); WeChat            & late-night social media engagement (11pm--1am) \\
c207 & 5\%   & YouTube (65\%); Zoom (7\%); Instagram (4\%); Gmail (2\%); WhatsApp (2\%)              & midday low-foreground hours \\
c291 & 27\%  & Netflix (28\%); Zoom (26\%); Coupang (8\%); WhatsApp (4\%); TikTok (4\%)              & evening streaming with brief messaging \\
c140 & 21\%  & Feishu (32\%); Instagram (14\%); WeChat (14\%); YouTube (14\%); Xiaohongshu (8\%)     & early-morning light foreground engagement (6--7am) \\
c425 & 80\%  & Crunchyroll (56\%); YouTube (11\%); Instagram (8\%); Xiaohongshu (5\%); WeChat (4\%)  & afternoon entertainment streaming \\
c442 & 79\%  & Zoom (55\%); Microsoft (25\%); Adobe (11\%); LinkedIn (1\%); YouTube (1\%)            & evening focused work session \\
c502 & 28\%  & Reddit (24\%); Medium (11\%); Spotify (8\%); YouTube (6\%); LinkedIn                  & evening article and forum browsing \\
c407 & 92\%  & Bilibili (25\%); Amazon Video (16\%); TikTok (12\%); Instagram (8\%); WeChat          & late-night entertainment streaming with messaging \\
\bottomrule
\end{tabular}
\end{table*}

\end{document}